\documentclass[lettersize,journal]{IEEEtran}
\usepackage{amsmath,amsfonts}
\usepackage{algorithmic}
\usepackage{algorithm}
\usepackage{array}
\usepackage[caption=false,font=normalsize,labelfont=sf,textfont=sf]{subfig}
\usepackage{textcomp}
\usepackage{stfloats}
\usepackage{url}
\usepackage{verbatim}
\usepackage{graphicx}
\usepackage{cite}

\usepackage{multirow}
\usepackage{textcomp}
\usepackage{stfloats}
\usepackage{url}
\usepackage{verbatim}
\usepackage{color,xcolor,soul}
\usepackage{booktabs}
\usepackage{color}
\usepackage{pifont}
\usepackage[switch]{lineno}
\usepackage{amsmath}
\usepackage{amssymb}
\usepackage{dsfont}
\usepackage{colortbl}  
\usepackage{xcolor}
\usepackage{array}   
\usepackage{soul}

\makeatletter
\let\myorg@bibitem\bibitem
\def\bibitem#1#2\par{%
	\@ifundefined{bibitem@#1}{%
		\myorg@bibitem{#1}#2\par
	}{%
		\begingroup
		\color{\csname bibitem@#1\endcsname}%
		\myorg@bibitem{#1}#2\par
		\endgroup
	}%
}

\makeatother
\usepackage{pifont}
\usepackage[breaklinks=true,colorlinks,bookmarks=false,citecolor=blue,linkcolor=blue]{hyperref}

\begin{document}

\title{Towards Generalized Few-Shot Open-Set Object Detection}
\author{Binyi Su, Hua Zhang, Jingzhi Li, and Zhong Zhou  
	\thanks{This work was supported in part by the
		National Key Research and Development Program of China under Grant
		2021YFB3100800, Natural Science Foundation of China under Grant 62272018 and 62072454, Beijing Natural Science Foundation under Grant 4202084, Basal Research Fund of Central Public Research Institute of China under Grant 20212701. (\textit{Corresponding author: Zhong Zhou.})
		
		B. Su is with the State Key Laboratory of Virtual Reality Technology and Systems, School of Computer Science and Engineering, Beihang University, Beijing 100191, China (e-mail: Subinyi@buaa.edu.cn).
		
		H. Zhang and J. Li are with Institute of Information Engineering, Chinese
		Academy of Sciences, Beijing 100195, China (e-mail: zhanghua@iie.ac.cn, lijingzhi@iie.ac.cn).
		
		Z. Zhou is with the State Key Laboratory of Virtual Reality Technology and Systems, School of Computer Science and Engineering, Beihang University, Beijing 100191, China, and also with Zhongguancun Laboratory, Beijing 100190, China (e-mail: zz@buaa.edu.cn).
}}

\markboth{Journal of \LaTeX\ Class Files,~Vol.~14, No.~8, August~2021}%
{Shell \MakeLowercase{\textit{et al.}}: A Sample Article Using IEEEtran.cls for IEEE Journals}


\maketitle

\begin{abstract}
Open-set object detection (OSOD) aims to detect the known categories and reject unknown objects in a dynamic world, which has achieved significant attention. However, previous approaches only consider this problem in data-abundant conditions, while neglecting the few-shot scenes. In this paper, we seek a solution for the generalized few-shot open-set object detection (G-FOOD), which aims to avoid detecting unknown classes as known classes with a high confidence score while maintaining the performance of few-shot detection. The main challenge for this task is that few training samples induce the model to overfit on the known classes, resulting in a poor open-set performance. We propose a new G-FOOD algorithm to tackle this issue, named \underline{F}ew-sh\underline{O}t \underline{O}pen-set \underline{D}etector (FOOD), which contains a novel class weight sparsification classifier (CWSC) and a novel unknown decoupling learner (UDL). To prevent over-fitting, CWSC randomly sparses parts of the normalized weights for the logit prediction of all classes, and then decreases the co-adaptability between the class and its neighbors. Alongside, UDL decouples training the unknown class and enables the model to form a compact unknown decision boundary. Thus, the unknown objects can be identified with a confidence probability without any threshold, prototype, or generation. We compare our method with several state-of-the-art OSOD methods in few-shot scenes and observe that our method improves the F-score of unknown classes by 4.80\%-9.08\% across all shots in VOC-COCO dataset settings \footnote[1]{The source code is available at \url{https://github.com/binyisu/food}}.
\end{abstract}

\begin{IEEEkeywords}
Generalized few-shot open-set object detection, class weight sparsification classifier, unknown decoupling learner.
\end{IEEEkeywords}

\section{Introduction}
\IEEEPARstart{O}{bject} detection is a fundamental task in the field of computer vision, which involves not only recognizing objects but also determining their precise locations by drawing bounding boxes around them.
With the assistance of deep learning, object detection has achieved remarkable progress. However, existing object detection models \cite{FasterRCNN,SSD, yolo} are under a strong assumption that there exist enough samples for all the categories, which is time-consuming and expensive to annotate instances for supervised training. 

To alleviate this issue, few-shot object detection (FSOD) methods \cite{TFA, FSCE, DCNet, CME, FSOD-UP, SRR-FSOD, Re-RCNN, DeFRCN, Pseudo-Labelling, FSL, FSRDD} are developed to reduce the data dependence of the CNN models. 
FSOD aims to train a detector based on a few samples. Various approaches have presented significant improvements in the FSOD problem. However, these methods hold a closed-set assumption, where the training and testing sets share the same classes. In open-set situations, there are countless unknown classes, not included in the training set. These unknown objects can easily disrupt the rhythm of the closed-set models, causing them to identify the unknown classes as known ones with a high confidence score \cite{OpenDet}. 

\begin{figure}[!t]
	\centering
	\includegraphics[width=8.8cm]{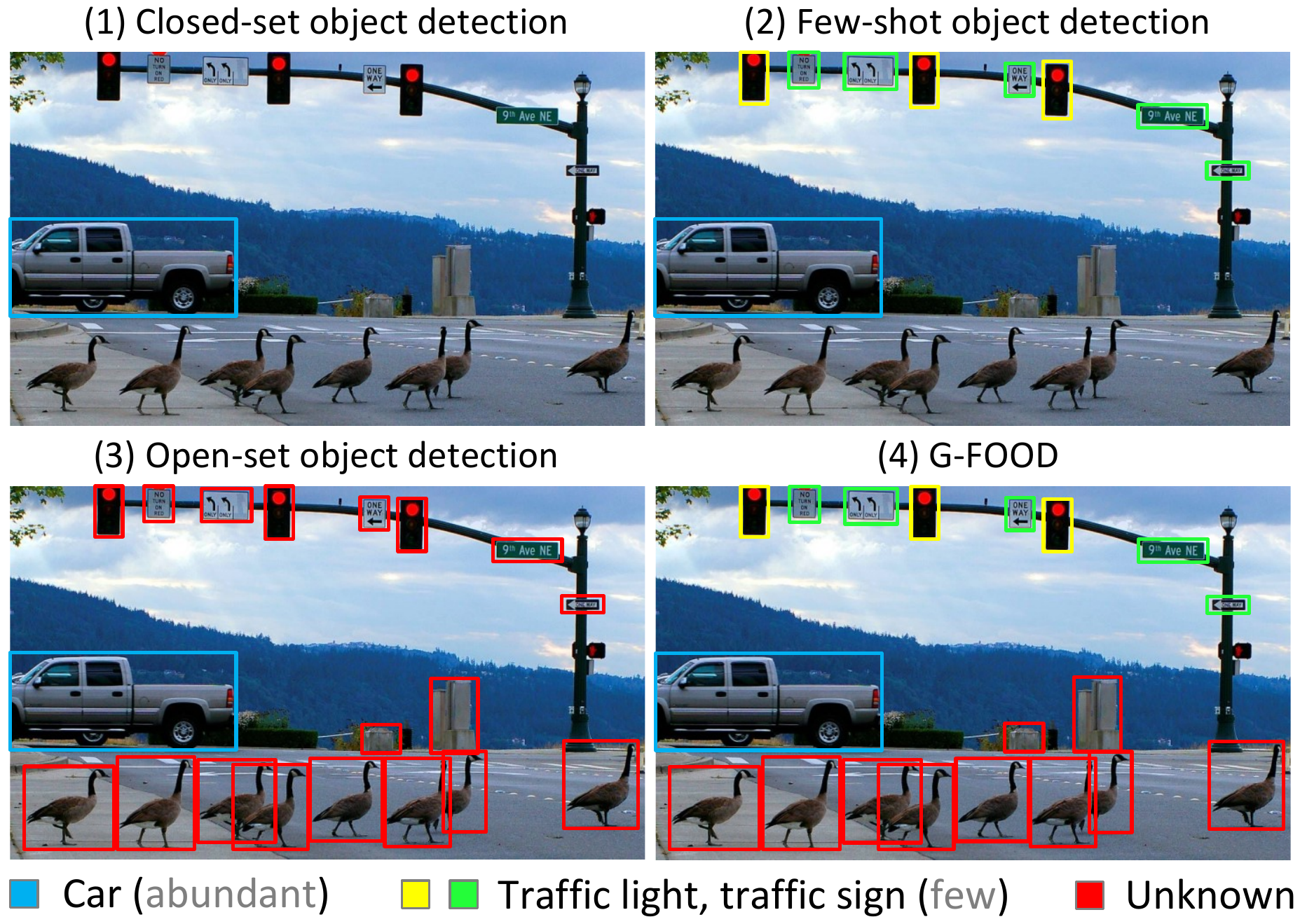}
	\caption{The visualization of different tasks: closed-set object detection (CSOD), few-shot object detection (FSOD), open-set object detection (OSOD), and generalized few-shot open-set object detection (G-FOOD). In CSOD and FSOD tasks, unknown objects are ignored or incorrectly classified into the set of known classes. The OSOD task can reject unknown class, but it usually requires data-abundant known classes for training \cite{FSOSR2021}. Our G-FOOD task can identify the data-abundant and data-hungry known objects while rejecting  unknown objects based on limited training data, which provides a better open-scene understanding paradigm.}\label{fig1}
\end{figure}


To make the model better handle the open-set scenarios, open-set object detection (OSOD) \cite{OS, DS, OpenDet} has been constantly investigated, where the detector trained on the closed-set datasets is asked to detect all known classes and reject unknown classes in open-set conditions. These OSOD methods leverage good representations of the known classes with sufficient training samples to construct unknown-aware detectors. However, open-set detectors suffer from a serious over-fitting problem \cite{FSOSR2021} with few known training samples, which greatly degrades the performance of open-set detection. 


In this paper, we seek a solution for the unexplored generalized few-shot open-set object detection (G-FOOD) problem, which commits to training a detector for unknown rejection using few samples. The concept of G-FOOD is expressed in Fig. \ref{fig1}. G-FOOD has enormous value in safety-critical applications such as autonomous driving and medical analysis. 
For example, in autonomous driving scenarios, encountering data-hungry known classes such as rare animals and unknown objects such as unexpected obstacles can potentially pose high risks to the safe operation of autonomous vehicles. Typically, the car needs to detect all known classes (including data-abundant classes and data-hungry classes) and reject the countless unknown objects as an ``unknown'' class in dynamic scenes. There are three irresistible reasons to solve the G-FOOD problem. First, an open-set detector that can identify the few-shot classes is more useful than the one that does not. Second, data-abundant open-set detection is a challenge in all settings. However, few-shot open-set detection is harder than the data-abundant open-set detection \cite{FSOSR2020}. With limited training data, there is a higher chance that the available few-shot examples may not fully represent the intra-class variations and complexities present in the known classes, causing serious overfitting problems. This can lead to poor generalization and difficulty in distinguishing known classes from unknown during inference. Third, like open-set detection, the main challenge of few-shot detection is to make accurate decisions for the unknown data during inference. Once the few-shot detectors have the ability for unknown rejection, they are
likely to be more robust for real-world applications. 




Actually, the dataset in real-world scenes often exhibits long-tail distributions \cite{LOCE} and there are always some unexpected categories that are not included in the dataset. G-FOOD aims to use this unbalanced dataset to train a detector that can identify all known classes and reject the countless unknown classes. 
G-FOOD could be viewed as an extension of the few-shot open-set recognition (FSOSR) \cite{FSOSR2020,FSOSR2021,R3CBAM, FSOSR2022}, which focuses on assigning a single label to an entire image without providing location information. However, our method is not to migrate the methods from FSOSR to G-FOOD. For example, the previous FSOSR studies adopt pseudo-unknown sample generation methods \cite{FSOSR2020,R3CBAM} or prototype-based methods \cite{FSOSR2021, FSOSR2022} to identify the unknown classes, however, our method is independent of the pseudo-unknown sample generation or the prototypes that denote the average feature of one class. Furthermore, the scarcity of training data induces the model to easily overfit on the few-shot known classes, making it challenging for the model to generalize and accurately reject instances of the unknown classes in real-world scenarios \cite{FSOSR2021}.
Therefore, how to solve the over-fitting problem without degrading the performance of few-shot known classes during unknown rejection becomes our main intention.


We know that dropout \cite{Dropout} suppresses over-fitting by reducing the co-adaptations between neurons. Thus, we draw inspiration from the consensus that reduction of the co-adaptability between the class and its neighbors can effectively suppress the over-fitting problem. We propose to reject the unknown classes by decoupling the co-adaptations between the known and unknown classes from two aspects: 1) The optimization process for the unknown class does not consider the interactions with the known classes, decoupling training it; 2) The classifier randomly sparses parts of the normalized weights for the class logit prediction, and then decreases the co-adaptability between the class and its neighbors.



To this end, we propose a novel generalized few-shot open-set object detector, named FOOD, which is threshold-free, prototype-free, and generation-free to reject unknown objects. This means that our approach does not rely on fixed thresholds or specific example representations of known classes, and avoids potential errors or biases arising from the pseudo-unknown sample generation process. 
Innovatively, FOOD employs Faster R-CNN \cite{FasterRCNN} as the base detector, where we replace the original classifier with a novel class weight sparsification classifier (CWSC) and additionally plug a novel unknown decoupling learner (UDL). These two modules cooperate with each other to solve the over-fitting problem of G-FOOD. After good optimization, our model can achieve the best G-FOOD performance than other state-of-the-art (SOTA) methods. Our method is also compatible with various backbones, including transformer-based architectures \cite{Swin-T}, which can leverage the advantages of different backbone models, such as their ability to capture long-range dependencies and semantic relationships.
The main contributions are threefold: 
\begin{itemize}
	\item We define a new problem, called generalized few-shot open-set object detection (G-FOOD), which aims to quickly train a detector based on a few labeled samples while rejecting all detected unknown objects.
	\item We propose a novel G-FOOD algorithm (FOOD) with two well-designed modules: CWSC and UDL, which can improve the model's generalization ability for unknown rejection in few-shot scenes.
	\item We developed the first G-FOOD benchmark. We modify several state-of-the-art OSOD methods into G-FOOD methods. Compared with these methods, our FOOD improves the F-score of unknown classes by 4.80\%-9.08\% across all shots in VOC-COCO dataset settings. 
\end{itemize}

This paper is organized as follows:  Section \uppercase\expandafter{\romannumeral2} shows an overview of the related works. Section \uppercase\expandafter{\romannumeral3} introduces the proposed method. Section \uppercase\expandafter{\romannumeral4} presents the extensive experiments. Finally, Section \uppercase\expandafter{\romannumeral5} concludes the paper.

\section{Related Work}
\subsection{Few-Shot Object Detection}
Since the conventional detectors based on supervised learning require abundant annotated samples for training, few-shot object detection (FSOD) has received significant progress recently \cite{TFA,DeFRCN,FSCE, DCNet, FSOD-UP,CME, Pseudo-Labelling,FSRW, MetaRCNN, FSOD, MetaDETR}. FSOD can be roughly divided into three types. 1) Meta learning-based methods. This type of works aims to learn the task-level knowledge that can be adapted to the new task with few support samples, such as FSRW \cite{FSRW}, Meta-RCNN \cite{MetaRCNN}, FSOD \cite{FSOD}, and Meta-DETR \cite{MetaDETR}.
2) Transfer-learning based approaches. This line of works adopts a simple two-stage fine-tuning strategy to train the detector, \emph{i.e.}, base-training and few-shot fine-tuning phases, which expects to transfer the general knowledge learned from the base-training phase to the few-shot fine-tuning phase, such as TFA \cite{TFA}, FSCE \cite{FSCE}, and DeFRCN \cite{DeFRCN}.
3) Pseudo-sample generation approaches. This form of works views FSOD as a data unbalanced problem, they employ the data-augmentation technologies to generate the samples of few-shot classes and train the detector end-to-end, such as \cite{Pseudo-Labelling}.



Although these FSOD methods focus on good detection performance under the closed-set settings, the unknown rejection capability is not guaranteed. We extend a simple FSOD method TFA \cite{TFA} with various OSOD methods to identify the unknown objects and find that our method can better reject the unknown classes in few-shot scenes than other SOTA methods.

\subsection{Open-Set Object Detection}
Open-set object detection (OSOD) methods intend to detect all known classes and reject the unknown classes, simultaneously. 
According to the acquisition way of the unknown samples, OSOD can be mainly divided into three categories. 1) Virtual unknown sample generation. This type of methods synthesizes the virtual unknown samples to train the unknown branch in the feature space \cite{VOS} or image space \cite{imgood}. 2) Select unknown samples from the background. This kind of works selects the background boxes with high uncertainty scores as the unknown class to train the open-set detector, such as ORE \cite{ORE}, UC-OWOD \cite{UCOWOD}, ROWOD \cite{ROWOD}, and OSODD \cite{OSODD}. 3) Select unknown samples from the known classes. This form of works chooses the known samples with high uncertainty scores as the unknown class to train the open-set detector, such as PROSER \cite{PROSER} and OpenDet \cite{OpenDet}.
Moreover, there exist several threshold-based methods. This type of works uses the energy or entropy as the uncertainty score, which is compared with a threshold to reject the unknown class, such as OS \cite{OS}, DS \cite{DS}, MCSSD \cite{MCSSD}, and GMM-Det \cite{GMMDet}. There are also several methods aiming at detecting all open-world objects without classifying them, such as OLN \cite{OLN} and LEDT \cite{LDET}.

The previous methods of OSOD usually need abundant samples of the known classes to train the model. However, this cannot be satisfied in the few-shot conditions, which causes a serious over-fitting problem of the model to the few-shot known classes, resulting in a poor open-set performance. Inspired by the dropout \cite{Dropout}, reduction of the neuron co-adaptations in optimization can efficiently suppress the over-fitting issue, we propose a class weight sparsification classifier and an unknown decoupling learner to dilute dependencies between all known and unknown classes.

\subsection{Few-Shot Open-Set Recognition}
Few-shot open-set recognition (FSOSR) has fascinated scant attention recently. However, to the best of our knowledge, generalized few-shot open-set object detection is still not exploited. Here, we present several FSOSR works. PEELER \cite{FSOSR2020} utilizes the pseudo-unseen class samples generated from seen classes to train the model. SnaTCHer \cite{FSOSR2021} measures the distance between the query and the transformed prototype, then a distance threshold is set to identify the unseen classes. R3CBAM \cite{R3CBAM} leverages the outlier calibration network to recognize the objects in FSOSR scenes. SEMAN-G \cite{FSOSR2022} learns an unseen prototype that automatically estimates a task-adaptive threshold for unseen rejection. Different from FSOSR, G-FOOD is indeed a more challenging task, because it involves not only identifying known and unknown object classes but also accurately localizing them in the image. Simultaneously, G-FOOD also has a background class, which often confuses the detector.

\subsection{Generalized few-shot open-set object detection} As illustrated in Table \ref{table_000}, we analyze the distinctions between our G-FOOD task and other related tasks including few-shot object detection (FSOD) \cite{TFA}, object discovery (OD) \cite{OD}, open-world object detection (OWOD) \cite{ORE}, out-of-distribution detection (OOD) \cite{VOS}, open-set object detection (OSOD) \cite{OpenDet}, open-set object detection and discovery (OSODD) \cite{OSODD}. The first indicator (few-shot training) refers to the training of deep learning models using a limited amount of labeled examples. The second (known detection) refers to the ability of a model to identify instances that belong to known classes. The third (unknown detection/rejection) denotes that the model could identify instances that do not belong to any of the known classes or reject instances deemed as unknown. The fourth (unknown discovery) denotes discovering the categories of unknown objects with multiple classes in an unsupervised manner \cite{OD}. The fifth  (incremental learning) aims to purposefully learn unknown objects in an incremental fashion \cite{ORE}. The sixth (cross-distribution evaluation) involves testing the model's performance on distributions that were not part of its training data \cite{VOS}. Our G-FOOD is characterized by few-shot training data, known and unknown detection, without unknown discovery and incremental learning, and not require cross-distribution evaluation, which is first explored in our paper.

\begin{table}[]
		\caption{Distinctions between different tasks.}\label{table_000}
	\resizebox{\linewidth}{!}{
	\begin{tabular}{|l|c|c|c|c|c|c|}
		\hline
		Task   & \begin{tabular}[c]{@{}c@{}}Few-shot\\ training\end{tabular} & \multicolumn{1}{c|}{\begin{tabular}[c]{@{}c@{}}Known\\ detection\end{tabular}} & \begin{tabular}[c]{@{}c@{}}Unknown \\ detection/ \\ rejection\end{tabular} & { \begin{tabular}[c]{@{}c@{}}Unknown\\ discovery\end{tabular}} & \begin{tabular}[c]{@{}c@{}}Incremental\\ learning\end{tabular} & \begin{tabular}[c]{@{}c@{}}Cross-\\ distribution\\ evaluation\end{tabular} \\ \hline
		FSOD \cite{TFA}   & Yes                                                         & Yes                                                                            & No                                                           & No                                                                                 & No                                                             & Not required                                                                  \\ \hline
		OD \cite{OD}    & No                                                          & No                                                                             & Yes                                                          & Yes                                                                                & No                                                             & Not required                                                                  \\ \hline
		OWOD \cite{ORE}   & No                                                          & Yes                                                                            & Yes                                                          & No                                                                                 & Yes                                                            & Not required                                                                  \\ \hline
		OOD \cite{VOS}   & No                                                          & Yes                                                                            & Yes                                                          & No                                                                                 & No                                                             & Required                                                                      \\ \hline
		OSOD \cite{OpenDet}  & No                                                          & Yes                                                                            & Yes                                                          & No                                                                                 & No                                                             & Not required                                                                  \\ \hline
		OSODD \cite{OSODD} & No                                                          & Yes                                                                             & Yes                                                          & Yes                                                                                & No                                                             & Not required                                                                  \\ \hline
		G-FOOD & Yes                                                         & Yes                                                                            & Yes                                                          & No                                                                                 & No                                                             & Not required                                                                  \\ \hline
\end{tabular}}
\end{table}


\section{Proposed Method}
\subsection{Problem Setup}
We define the problem setup with reference to TFA \cite{TFA} and OpenDet \cite{OpenDet}. We are given an object detection dataset $D=\{(x,y),x\in \textbf{X},y\in \textbf{Y}\}$, where $x$ denotes an input image and $y={\{(c_i,b_i)\}}_{i=1}^I$ represents the objects with its class $c$ and its box annotation $b$. 
The dataset $D$ is divided into the training set $D_{tr}$ and the testing set $D_{te}$. $D_{tr}=D_B\cup D_N$ contains $K$ known classes $C_K=C_B\cup C_N=\{1,...,K=B+N\}$, where $C_B=\{1,...,B\}$ expresses $B$ data-abundant base classes, and $C_N=\{B+1,...,K\}$ denotes $N$ data-hungry novel classes, each with $M$-shot support samples. Here, novel classes express few-shot known classes rather than unknown classes defined in open-set communities \cite{DS,OpenDet}. $D_B$ and $D_N$ denote the training instances of the base and novel classes, respectively. We test the detector in $D_{te}$ that includes $C_K=C_B\cup C_N$ known classes and $C_U$ unknown classes. Duo to the countless unknown categories, we merge all of them into one class $C_U=\{K+1\}$. Our goal is to employ the unbalanced data $D_{tr}=D_B\cup D_N$ to train a detector, which can be used to identify the base classes $C_B$, the novel classes $C_N$, the unknown class $C_U$, and the background class $C_{bg}$. Here, regions that do not contain any annotated objects of interest effectively serve as training samples for the background class. This is pivotal for enabling the model to explicitly identify regions that do not align with any specific object category (base, novel, or unknown).

\begin{figure*}[!t]
	\centering
	\includegraphics[width=17.7cm]{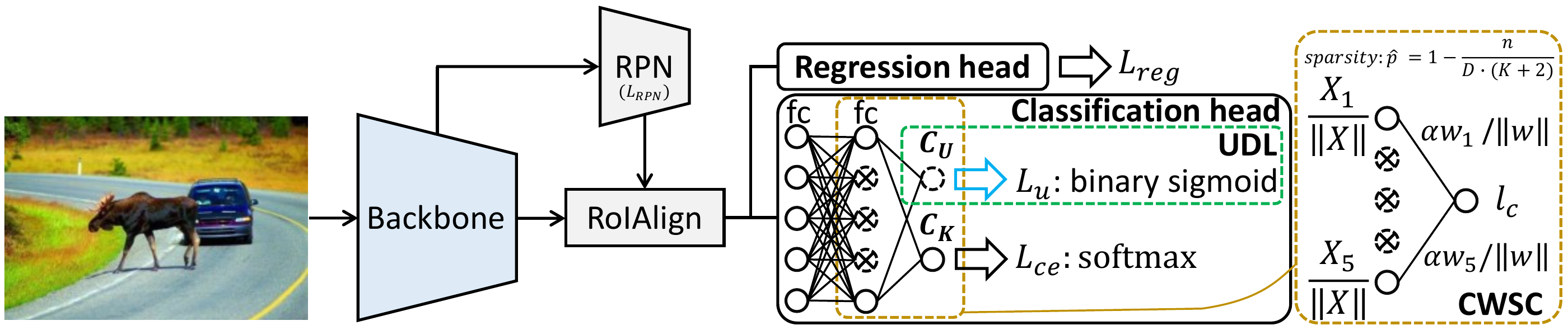}
	\caption{The framework of our FOOD for generalized few-shot open-set object detection. 
		Compared to the standard Faster R-CNN, FOOD plugs a novel class weight sparsification classifier (CWSC) and a novel unknown decoupling learner (UDL).  We sparsity the normalized weights for the class logit prediction and simultaneously optimize a binary sigmoid classifier and a multiply softmax classifier in the classification head. Our method is characterized by no pseudo-unknown sample generation, prototype-free, and threshold-free to reject unknowns in few-shot scenes.
	}\label{fig2}
\end{figure*}

\subsection{Baseline Setup}
As shown in Fig. \ref{fig2}, Faster R-CNN \cite{FasterRCNN} is adopted as the base detector that is composed of a backbone, a region proposal network (RPN), a region of interest alignment (RoIAlign) layer, plus two heads (a regression head \cite{FasterRCNN} and a classification head). Compared with the standard Faster R-CNN, three tricks are utilized to improve the detector.  (1) \textbf{Classification and regression decoupling}: The original two heads contain two shared fully connected (fc) layers and two separate fc layers for classification and regression. In order to prevent the classification task from disturbing the regression task, the shared fc layers are replaced by two parallel fc layers. Simultaneously, the class-specific box regression is changed to class-agnostic. For example, the standard output of the box predictor (box regression head) is $4\times (K+2)$, now we set it as 4, where $K+2$ denotes $K$ known classes, 1 unknown class, and 1 background class.  It means that for each region proposal, we predict one box for all classes, instead of one box
per class. The above operations are utilized to decouple the classification and the regression, and then provide convenience to tackle the generalized few-shot open-set object detection task.  

(2) \textbf{Two-stage fine-tuning strategy}: 
Following TFA \cite{TFA}, the training process consists of a base training stage and a few-shot fine-tuning stage. In the base training stage, we employ abundant samples of the base classes $C_B$ to train the entire base detector from scratch, such as Faster R-CNN. Then, in the few-shot fine-tuning stage, we create a small balanced training set with $M$ shots per class, containing both base and novel classes. The balanced dataset is first used to train the last linear layers of the base detector while freezing the other parameters of the model (linear probing \cite{ICLR2022oral}), and then fine-tune all the parameters of the model (fine-tuning) in a soft-freezing way \cite{DeFRCN}, which employs a scaled gradient to slowly update the parameters of the backbone network to get the generalized few-shot open-set object detector. Note that there are two layers for linear probing including the last box classification layer and the last box regression layer, namely the last linear layers of the base detector.

(3) \textbf{Classifier placeholder}: Classifier placeholders \cite{PROSER} refer to variables or symbols that represent the preserved classes in the classification head. In the base training stage, we reserve the dummy classifier placeholders for the novel classes $C_N$ and an unknown class $C_U$ to augment the class number of the open-set classifier. These placeholders reserved for the novel classes will be optimized in the few-shot fine-tuning stage. Overall, there are two advantages for the predefined classifier placeholder: one is that the classifier placeholder omits the additional model surgery step \cite{TFA, DeFRCN}, which is used to augment the number of model categories from the base training stage (base classes $C_B$) to the few-shot fine-tuning stage (base classes $C_B$ + novel classes $C_N$) for the closed-set classifier. This means that our method simplifies the training process of the FSOD methods based on transfer learning. Another is that the dummy sub-classifier for the unknown class is necessary to optimize our proposed unknown decoupling learner (UDL), which can reject the unknown objects without relying on the pseudo-unknown samples for training.


\subsection{Class Weight Sparsification Classifier (CWSC)}
Few training samples induce the model to overfit on the known classes, thus the model cannot extract the generalization features that can be used for the unknown rejection. The neuron dropout theory \cite{Dropout} has been employed to suppress the over-fitting issue. The dropout randomly drops some weights during training. During the prediction process, if we still randomly drop some weights, the model will typically produce unstable results and hurt the detection performance.
However, suppose all weights are reserved for prediction at test time. In that case, it will result in a large expected (or mean) difference in the output between training (with dropout) and testing (without dropout), leading to performance degradation. So we need to make the output expected value of training and testing as consistent as possible. Our CWSC can be understood as class weight sparsification for unknown rejection in few-shot scenes. As shown in Fig. \ref{fig2}, our CWSC sparses the normalized weights $w/\vert\vert w\vert\vert$ rather than $w$ and measures the cosine similarity with the normalized feature $X/\vert\vert X\vert\vert$ for the class logit prediction, where $\vert\vert \cdot\vert\vert$ denotes the L2 norm. This makes the expected difference with and without sparsification become bounded, and then the model is more stable and prominent for open-set detection in few-shot scenes.


Specifically, $w\in\mathbb{R}^{D\times (K+2)}$ represents the weights of the last linear mapping to $(K+2)$ classes, where $D$ denotes the weight dimension for each class. We use the sparsity probability $\widehat p=1-\frac n{D\cdot(K+2)}$, where $n$ denotes the number of randomly selected weights for the class logit prediction. Then, the retention probability of the weight can be denoted as $p=1-\widehat p$. The weights can be denoted as $R*w/\vert\vert w\vert\vert$, where $R\in{\{0,1\}}^{D\times (K+2)}$ is initialized with $R_{i,j}\sim\mathrm{Bernoulli}(p)$ and $\ast$ represents the element-wise product. The total number of 1 in $R$ is $n$, where 1 denotes retaining the weight and 0 denotes dropping the weight. The class logit output could be expressed as:  
\begin{equation}
	l_c=\alpha\frac X{\vert\vert X\vert\vert}\left[R\ast\frac w{\vert\vert w\vert\vert}\right],
\end{equation} 
where $\alpha$ denotes a positive temperature factor. In our CWSC, if we set the retention probability $p$, the expected value (or mean) of the logit output $l_c$ with sparsification falls in $\lbrack-p\alpha, p\alpha\rbrack$, without sparsification, it falls in $\lbrack-\alpha, \alpha \rbrack$, thus the expected difference with and without sparsification is $\mu_{diff}^{cos}\in\lbrack-(1+p)\alpha, (1+p)\alpha\rbrack$. For the conventional output $\widehat l_c=Xw$, the expected difference of the logit output with and without sparsification is $\mu_{diff}\in(-\infty,+\infty)$. Compared with $\mu_{diff}$, $\mu_{diff}^{cos}$ is bounded, which means the output distribution of our CWSC with and without sparsification is more consistent, producing better results for G-FOOD.

Next, we theoretically analyze why the CWSC can suppress the over-fitting issue. Assuming a linear regression task, $y\in\mathbb{R}^{N}$ is the ground truth label, the model tries to find a $w\in\mathbb{R}^{D}$ to minimize
\begin{equation}\label{eq1}
	\vert\vert y-\alpha \frac X{\vert\vert X\vert\vert}\frac w{\vert\vert w\vert\vert}\vert\vert^{2}.
\end{equation}
We set $\overset-x=X/\vert\vert X\vert\vert$ and $\overset-w=w/\vert\vert w\vert\vert$. When the weight sparsification is adopted, the objective function becomes 
\begin{equation}\label{eq2}
	\underset w{minimize}\;{\mathbb{E}}_{R\sim\mathrm{Bernoulli}(p)}\lbrack{\vert\vert y-\alpha \overset-x(R\ast\overset-w)\vert\vert}^2\rbrack.
\end{equation}
This can reduce to 
\begin{equation}\label{eq3}
	\underset w{minimize}{\vert\vert y-\alpha p\overset-x\overset-w\vert\vert}^2+\alpha^2p(1-p){\vert\vert\tau\overset-w\vert\vert}^2, 
\end{equation}
where $\tau=({diag(\overset-x^{\mathrm T}\overset-x)}{)}^{1/2}$. We set $\widetilde w=\alpha p\overset-w$, then 
\begin{equation}\label{eq4}
	\underset w{minimize}{\vert\vert y-\overset-x\widetilde w\vert\vert}^2+{\textstyle\frac{(1-p)}p}{\vert\vert\tau\widetilde w\vert\vert}^2.
\end{equation}
Eq. \ref{eq4} can be viewed as a ridge regression with a particular form for $\tau$. If a particular data dimension changes a lot, the regularizer tries to sparse its weight more \cite{Dropout}, and then the over-fitting problem is alleviated through regularization. We can control the strength of the regularization by adjusting the sparsity probability $\widehat p=1-p$. For example, if we set $\widehat p=0$, then $p=1$, the regularization term is 0, which means that the regularizer does not work. As the sparsity probability $\widehat p$ increases, the regularization constant grows larger and the regularization effect becomes more pronounced. Our method benefits from the over-fitting suppression, and we verify that the class weight sparsification greatly improves the model's generalization ability for the open-set detection in few-shot scenes. A derivation of Eq. \ref{eq1}-\ref{eq4} is presented in \nameref{appendix}.

\subsection{Unknown Decoupling Learner (UDL)}
Unknown decoupling learner (UDL) plays a decisive role in rejecting the unknown class, which provides a dummy unknown class for decoupling optimization, and then boosts the model to form a compact unknown decision boundary. The UDL does not depend on the pseudo-unknown samples generation method \cite{FSOSR2020} to train the dummy unknown class, because the data distribution of the unknown class is more complex and changeable, the generated fake unknown samples often fail to simulate the real distribution of the unknown data. Inspired by the fact that the known class data and the unknown class data are often orthogonal \cite{ICLR2022oral}, we propose to decouple optimizing the unknown class without relying on the predictions of the known classes. Thus, we select a sigmoid function that normalizes the predicted unknown logit to estimate the unknown probability: 
\begin{equation}\label{eq5}
	p_u(l_{C_U})=\frac1{1+e^{-\delta\cdot l_{C_U}}},
\end{equation}
where $l_{C_U}$ represents the predicted logit for the dummy unknown class $C_U$ in the classification branch and $\delta$ is used to adjust the slope of the sigmoid function. Why do we choose the sigmoid function to compute the unknown probability $p_u$ instead of softmax? If we select the softmax, the objective function for the unknown branch becomes: 
\begin{equation}\label{eq7}
\begin{split}
L_u\downarrow=-\log p_u\textcolor{blue}{\uparrow}=-\log\frac{e^{l_{C_U}\textcolor{blue}{\uparrow}}}{{\displaystyle\sum_{c_i\in C,c_i\neq c_\ast}}e^{l_{c_i}}+e^{l_{c_\ast}\textcolor{blue}{\uparrow}}}
\end{split}
\end{equation}
where $C=C_B\cup C_N\cup C_U \cup C_{bg}$, $\textcolor{blue}{\uparrow}$ denotes that the optimized target should be as large as possible in optimization and $\textcolor{blue}{\downarrow}$ represents the opposite. There are no real unknown samples or groundtruths in the training process. The positive pseudo-unknown samples used in this method are high conditional energy (see Eq. \ref{eq8}) samples selected from known classes or backgrounds. In addition to the groundtruth of an unknown class, each pseudo-unknown sample also has its own real known class or background label. During the unknown optimization, the model may predict high logits ($l_{C_U}$ and $l_{c_\ast}$) for the unknown class $C_U$ and its real ground-truth class $c_\ast$, simultaneously, which causes optimization conflicts, making it difficult for the unknown class to converge. While the unknown probability product by the sigmoid function does not care about the logit outputs for other classes (see Eq. \ref{eq5}), we call it unknown decouple training, which can alleviate the above problem.


Therefore, the loss function of the dummy unknown class (unknown decoupling loss) is defined as follows:

\begin{equation}\label{eq_3}
	\begin{split}
		\mathcal{L}_{u}=\frac1{N_{pos}}\sum_{i=1}^{N_{pos}}\log (1+e^{-\delta_1\cdot l_{C_U}(\mathcal{S}_{pos}^i)})\\+\frac1{N_{neg}}\sum_{j=1}^{N_{neg}}\log(1+e^{-\delta_2\cdot l_{C_U}(\mathcal{S}_{neg}^j)}),
	\end{split}
\end{equation}
where $N_{pos}$ and $N_{neg}$ represent the number of positive pseudo-unknown samples $\mathcal{S}_{pos}$ and $\mathcal{S}_{neg}$ selected from foreground samples of known classes and background samples, respectively.
Note that negative pseudo-unknown samples are not required to optimize the sigmoid-based classifier because it assigns a binary label (0 or 1) to each proposal, indicating the presence or absence of an object in each proposal \cite{yolo}.
How do we select the positive pseudo-unknown samples ($\mathcal{S}_{pos}$ and $\mathcal{S}_{neg}$) from all foreground proposals $\mathcal{S}_{fg}$ and background proposals $\mathcal{S}_{bg}$ to train the UDL branch? Ideally, a model should learn a more compact decision boundary that produces low uncertainty for the known data, with high uncertainty for unknown data elsewhere. Fortunately, the energy score \cite{Energy,VOS} has been used to measure the uncertainty of the object, where the larger the energy, the higher the uncertainty of the sample \cite{VOS}. The motivation for selecting high-energy regions as pseudo-unknown samples is rooted in the assumption that objects from unknown classes are not seen during training. Since the model does not have specific knowledge about those unknown classes, unknown classes may have low-logit predictions compared with known and background classes frequently seen during training. The lower the logit, the higher the energy (see Eq. \ref{eq8}). Here, we propose to select the foreground and background samples with high conditional energy as positive pseudo-unknown samples to train the UDL branch. For a region proposal $\mathcal{S}_j\in \mathcal{S}_{fg}\cup \mathcal{S}_{bg}$, the conditional energy score is defined as:
\begin{equation}\label{eq8}
	E(\mathcal{S}_j)_{c_i\neq C_U}=-\log\sum_{c_i\neq C_U,c_i\in C}e^{l_{c_i}(\mathcal{S}_j)},
\end{equation}
where $l_{c_i}(\cdot)$ is the predicted logit for class $c_i$. Since there are no real unknown class training samples, the term $exp(l_{C_U}(\mathcal{S}_j))$ will become an interference term in the energy-based sampling process. Thus, we discard it and select the top-$k$ samples ranked by the conditional energy score $E(\mathcal{S}_j)_{c_i\neq C_U}$ to optimize the unknown decoupling loss $L_u$. The positive pseudo-unknown samples ($\mathcal{S}_{pos}$ and $\mathcal{S}_{neg}$) used to train the UDL branch can be denoted as:
\begin{equation}\label{eq5_1}
	\mathcal{S}_{pos}=\underset{\mathcal{S}_{j'}\in \mathcal{S}_{fg}}{topk}\left(-\log\sum_{c_i\neq C_U,c_i\in C}e^{l_{c_i}(\mathcal{S}_{j'})}\right)_{k=N_{pos}},
\end{equation}
\begin{equation}\label{eq5_2}
	\mathcal{S}_{neg}=\underset{\mathcal{S}_{j''}\in \mathcal{S}_{bg}}{topk}\left(-\log\sum_{c_i\neq C_U,c_i\in C}e^{l_{c_i}(\mathcal{S}_{j''})}\right)_{k=N_{neg}}.
\end{equation}

By selecting high conditional energy proposals as pseudo-unknown samples for training, the model can focus on learning to treat high conditional energy samples as unknown classes and low conditional energy samples as known classes or the background, thus assisting in forming the unknown decision boundary of the model.
Overall, the sub-classifier of the dummy unknown class in UDL is equivalent to merging the binary sigmoid classifier into the multiply softmax classifier of real known classes. Then, the model can synchronously identify the specific class of known and unknown objects in inference. Instead of asynchronously distinguishing the unknown class from the known classes like the threshold-based methods \cite{DS,MCSSD}, if it is not an unknown class, the model would distinguish its specific known class.

\subsection{Overall Optimization}
With the unknown decoupling loss $L_u$, the final loss function is defined as follows:
\begin{equation}\label{eq9}
	\mathcal{L}=\mathcal{L}_{RPN}+\mathcal{L}_{reg}+\mathcal{L}_{ce}(D_B\cup D_N)+\lambda \mathcal{L}_u(\mathcal{S}_{pos}\cup \mathcal{S}_{neg}),
\end{equation}
where $\mathcal{L}_{RPN}$ is the RPN loss that consists of a binary cross-entropy loss and a regression loss. $\mathcal{L}_{reg}$ denotes the smooth L1 loss. $\mathcal{L}_{ce}$ expresses the cross-entropy loss. $\lambda$ is used to balance the proportion of the unknown decoupling loss $\mathcal{L}_u$.
The above four sub-losses can be used to jointly optimize the G-FOOD model $f_\theta$ as:
\begin{equation}
	\begin{split}
		\underset\theta{argmin(}{\mathbb{E}}_{(x,y)\in D_B\cup D_N}\mathcal{L}_{RPN,reg,ce}(x,y;f_\theta)\\+{\mathbb{E}}_{(x,y)\in \mathcal{S}_{pos}\cup \mathcal{S}_{neg}}\mathcal{L}_u(x,y;f_\theta)),
	\end{split}
\end{equation}
where $\theta$ denotes the learned weights for the G-FOOD model. 

\subsection{Inference}\label{inference}
During inference, we normalize the logits of all classes (base classes, novel classes, the unknown class, and the background class) by the softmax function to get the final classification score:

\begin{equation}\label{eq10}
	p_{c_{m}}=\underset{c_j\in C}{max}(\frac{e^{l_{c_j}}}{{\displaystyle\sum_{c_i\in C=C_B\cup C_N\cup C_U\cup C_{bg}}}e^{l_{c_i}}}).
\end{equation}
The specific class of the predicted box is $c_m$.
We not continue to choose sigmoid to compute the probability of the unknown class $p_{u}$ in inference. The reason is that if we choose sigmoid to compute $p_{u}$ at test time, our method would become a threshold-based method to divide the unknown class, which means we need to set a threshold to identify the unknown class. However, softmax is a threshold-free method during testing, the class with the maximum softmax probability (MSP) is the final predicted class of the box. Thus, we select it and our method becomes threshold-free for unknown rejection based on a few training samples.

\begin{table*}[]
	\caption{The Generalized Few-Shot Open-Set Object Detection Results on \textbf{VOC10-5-5} Dataset Settings. $\uparrow$ Indicates That the Larger the Evaluation Metrics, the Better the Performance. \textbf{Bold} Numbers denote Superior Results. For a Fair Comparison, We Report the Average Results of 10 Random Runs for All Comparison Methods.}\label{table_0}
	\renewcommand\arraystretch{1.05}
	\resizebox{\linewidth}{!}{
		\begin{tabular}{l|ccccccc|ccccccc}
			\toprule[1pt]
			\multirow{2}{*}{\textbf{VOC10-5-5}} & \multicolumn{7}{c|}{1-shot}                                                                                        & \multicolumn{7}{c}{2-shot}                                                                                         \\
			& $mAP_K\uparrow$        & $mAP_B\uparrow$        & $mAP_N\uparrow$        & $AP_U\uparrow$        & $P_U\uparrow$         & $R_U\uparrow$          & $F_U\uparrow$          & $mAP_K$        & $mAP_B$        & $mAP_N$        & $AP_U$        & $P_U$         & $R_U$          & $F_U$          \\ \midrule[0.8pt]
			TFA\cite{TFA}                        & \textbf{45.31} & \textbf{63.71} & 8.50           & 0.00          & 0.00          & 0.00           & 0.00           & \textbf{46.48} & 62.55          & 14.34          & 0.00          & 0.00          & 0.00           & 0.00           \\
			DS+TFA\cite{DS}                    & 43.82          & 62.11          & 7.22           & 1.90          & \textbf{7.61} & 23.99          & 23.49          & 46.28          & \textbf{63.20} & 12.44          & 2.08          & \textbf{8.20} & 24.56          & 24.08          \\
			ORE+TFA\cite{ORE} & 43.25 	&61.31 	&8.62 	&1.23 	&5.97 	&18.25 	&17.89 	&44.86 	&61.08 	&12.17 	&1.58 	&5.78 	&22.19 	&21.58 \\
			PROSER+TFA\cite{PROSER}                 & 41.64          & 58.22          & 8.49           & 3.26          & 5.60          & 30.95          & 29.62          & 42.70          & 57.27          & 13.56          & 3.42          & 5.35          & 31.53          & 30.07          \\
			OpenDet+TFA\cite{OpenDet}                & 43.45          & 61.04          & 8.27           & 3.44          & 6.62          & 33.64          & 32.33          & 45.67          & 62.74          & 11.53          & 3.28          & 6.85          & 30.95          & 29.91          \\
			Our FOOD                   & 43.97          & 61.48          & \textbf{8.95}  & \textbf{4.26} & 3.58          & \textbf{43.72} & \textbf{39.35} & 45.85          & 61.37          & \textbf{14.80} & \textbf{4.61} & 3.51          & \textbf{45.52} & \textbf{40.70} \\ \midrule[0.8pt]
			\multirow{2}{*}{VOC10-5-5} & \multicolumn{7}{c|}{3-shot}                                                                                        & \multicolumn{7}{c}{5-shot}                                                                                         \\
			& $mAP_K$        & $mAP_B$        & $mAP_N$        & $AP_U$        & $P_U$         & $R_U$          & $F_U$          & $mAP_K$        & $mAP_B$        & $mAP_N$        & $AP_U$        & $P_U$         & $R_U$          & $F_U$         \\ \midrule[0.8pt]
			TFA\cite{TFA}                        & 47.55          & 63.72          & 15.23          & 0.00          & 0.00          & 0.00           & 0.00           & 47.88          & 61.92          & 19.74          & 0.00          & 0.00          & 0.00           & 0.00           \\
			DS+TFA\cite{DS}                     & 46.89          & 63.09          & 14.48          & 2.11          & \textbf{8.32} & 23.62          & 23.20          & 48.01          & 62.38          & 19.27          & 1.91          & \textbf{8.85} & 19.99          & 19.74          \\
			ORE+TFA\cite{ORE} & 45.88 	&62.23 	&14.52 &	1.71 &	5.96 	&22.23 &	21.64 	&46.29 	&62.71 	&18.49 	&1.82 	&5.98 	&23.01 	&22.38 \\
			PROSER+TFA\cite{PROSER}                 & 43.30          & 57.30          & 15.16          & 3.23          & 5.32          & 32.30          & 30.76          & 45.12          & 57.64          & 20.08          & 3.56          & 5.34          & 32.68          & 31.10          \\
			OpenDet+TFA\cite{OpenDet}                & 46.47          & 62.66          & 14.09          & 3.15          & 6.47          & 30.62          & 29.53          & 47.56          & 62.39          & 17.90          & 3.41          & 6.89          & 32.13          & 31.01          \\
			Our FOOD                   & \textbf{48.48} & \textbf{64.30} & \textbf{16.83} & \textbf{4.50} & 3.54          & \textbf{44.52} & \textbf{39.94} & \textbf{50.18} & \textbf{63.72} & \textbf{23.10} & \textbf{4.63} & 3.51          & \textbf{45.65} & \textbf{40.80} \\ \midrule[0.8pt]
			\multirow{2}{*}{VOC10-5-5} & \multicolumn{7}{c|}{10-shot}                                                                                       & \multicolumn{7}{c}{30-shot}                                                                                        \\
			& $mAP_K$        & $mAP_B$        & $mAP_N$        & $AP_U$        & $P_U$         & $R_U$          & $F_U$          & $mAP_K$        & $mAP_B$        & $mAP_N$        & $AP_U$        & $P_U$         & $R_U$          & $F_U$          \\ \midrule[0.8pt]
			TFA\cite{TFA}                        & 51.10          & 63.56          & 26.19          & 0.00          & 0.00          & 0.00           & 0.00           & 56.20          & 65.68          & 37.23          & 0.00          & 0.00          & 0.00           & 0.00           \\
			DS+TFA\cite{DS}                     & 48.01          & 62.38          & 25.66          & 1.91          & \textbf{8.85} & 19.99          & 19.74          & 56.60          & 66.27          & 36.96          & 1.90          & \textbf{9.95} & 18.02          & 17.88          \\
			ORE+TFA\cite{ORE} & 48.17 	&62.18 	&25.40 &	1.84 &	6.02 	&23.48 	&22.82 &	56.41 &	65.83 &	35.84 &	1.88 &	6.12 &	23.77 &	23.11 \\
			PROSER+TFA\cite{PROSER}                 & 48.35          & 59.96          & 25.13          & 3.59          & 5.16          & 32.61          & 30.98          & 53.93          & 62.96          & 35.86          & 3.46          & 4.76          & 33.93          & 31.99          \\
			OpenDet+TFA\cite{OpenDet}                & 50.95          & 63.85          & 25.14          & 4.06          & 6.61          & 36.30          & 34.75          & 56.11          & 66.05          & 36.24          & 4.03          & 5.62          & 40.51          & 38.16          \\
			Our FOOD                   & \textbf{53.23} & \textbf{65.54} & \textbf{28.60} & \textbf{4.71} & 3.59          & \textbf{45.84} & \textbf{41.06} & \textbf{58.59} & \textbf{67.73} & \textbf{40.29} & \textbf{4.90} & 3.62          & \textbf{46.40} & \textbf{41.54} \\ \bottomrule[1pt]
	\end{tabular}}
\end{table*}

\section{Experiment}
\subsection{Experimental Setup}
\subsubsection{Datasets}
We construct the G-FOOD benchmarks using PASCAL VOC 2007+2012 \cite{VOC}, MS COCO 2017 \cite{COCO}, and LVIS \cite{LVIS}. There are two types of benchmark settings. One is the single-dataset benchmark: VOC10-5-5 and LVIS315-454-461, which means only one dataset (PASCAL VOC or LVIS) is used to construct the G-FOOD benchmark. Another is the cross-dataset benchmark: VOC-COCO, which means two datasets (PASCAL VOC and MS COCO) are used to construct the G-FOOD benchmark. The VOC-COCO setting can ensure that the model barely sees the unknown classes during training.

\textbf{VOC10-5-5:} The 20 classes of PASCAL VOC are divided into 10 base known classes $C_B$, 5 novel known classes $C_N$, and 5 unknown classes $C_U$ to evaluate the G-FOOD performance of our method. The novel known
classes $C_N$ have $M=$1, 2, 3, 5, 10, and 30 objects per class sampled
from the training data of PASCAL VOC. Here we select the test set of PASCAL  VOC 2007 for the generalized few-shot open-set evaluation. In this paper, we define the above dataset divisions as the \textbf{VOC10-5-5} settings, where $C_B$=\{aeroplane, bicycle, bird, boat, bottle, bus, car, cat, chair, cow\},
$C_N$=\{diningtable, dog, horse, motorbike, person\},
$C_U$=\{pottedplant, sheep, sofa, train, tvmonitor\}=\{unknown\}.

\textbf{VOC-COCO:} The MS COCO dataset contains 80 classes, 20 of which overlap with the PASCAL VOC dataset. We use 20 classes of PASCAL VOC and 20 non-VOC classes of MS COCO as the closed-set training data, where PASCAL VOC servers as the base known classes $C_B$ and the 20 non-VOC classes of MS COCO are the few-shot splits of novel known classes  $C_N$. The novel known
classes $C_N$ have $M=$1, 2, 3, 5, 10, and 30 objects per class sampled
from the training data of MS COCO. The remaining 40 classes of MS COCO are used as the unknown classes $C_U$, which are challenging. Meanwhile, we use the validation set of MS COCO for the generalized few-shot open-set evaluation. In this paper, we define the above dataset divisions as the \textbf{VOC-COCO} settings, where
$C_B$=\{aeroplane, bicycle, bird, boat, bottle, bus, car, cat, chair, cow, diningtable, dog, horse, motorbike, person, pottedplant, sheep, sofa, train, tvmonitor\},
$C_N$=\{truck, traffic light, fire hydrant, stop sign, parking meter, bench, elephant, bear, zebra, giraffe, backpack, umbrella, handbag, tie, suitcase, microwave, oven, toaster, sink, refrigerator\},
$C_U$=\{frisbee, skis, snowboard, sports ball, kite, baseball bat, baseball glove, skateboard, surfboard, tennis racket, banana, apple, sandwich, orange, broccoli, carrot, hot dog, pizza, donut, cake, bed, toilet, laptop, mouse, remote, keyboard, cell phone, book, clock, vase, scissors, teddy bear, hair drier, toothbrush, wine glass, cup, fork, knife, spoon, bowl\}=\{unknown\}.

\textbf{LVIS315-454-461:} LVIS dataset has a natural long-tail distribution, the classes of which are divided into 315 frequent classes (appearing in more than 100 images), 461 common classes (10-100 images), and 454 rare classes (less than 10 images). Here, we use frequent classes, rare classes, and common classes as base known classes, novel known classes, and unknown classes, respectively. It does not need the manual $M$-shot split, where the rare classes can directly serve as the few-shot training data. Meanwhile, we use the validation set of LVIS for the generalized few-shot open-set evaluation.

\begin{table*}[]
	\centering
	\caption{The Generalized Few-Shot Open-Set Object Detection Results on \textbf{VOC-COCO} Dataset Settings. $\uparrow$ Indicates That the Larger the Evaluation Metrics, the Better the Performance. \textbf{Bold} Numbers denote Superior Results. For a Fair Comparison, We Report the Average Results of 10 Random Runs for All Comparison Methods.}\label{table_1}
	\renewcommand\arraystretch{1.05}
	\resizebox{\linewidth}{!}{
		\begin{tabular}{l|ccccccc|ccccccc}
			\toprule[1pt]
			\multirow{2}{*}{\textbf{VOC-COCO}} & \multicolumn{7}{c|}{1-shot}                                                                                        & \multicolumn{7}{c}{2-shot}                                                                                          \\
			& $mAP_K\uparrow$        & $mAP_B\uparrow$        & $mAP_N\uparrow$       & $AP_U\uparrow$        & $P_U\uparrow$          & $R_U\uparrow$          & $F_U\uparrow$          & $mAP_K$        & $mAP_B$        & $mAP_N$        & $AP_U$        & $P_U$          & $R_U$          & $F_U$          \\ \midrule[0.8pt]
			TFA\cite{TFA}                                & 15.77          & 29.03          & \textbf{2.50} & 0.00          & 0.00           & 0.00           & 0.00           & 15.82          & 28.01          & \textbf{3.63}  & 0.00          & 0.00           & 0.00           & 0.00           \\
			DS+TFA\cite{DS}                             & 15.47          & 28.84          & 2.11          & 0.48          & \textbf{11.53} & 3.57           & 3.59           & 16.28          & 29.36          & 3.21           & 0.51          & \textbf{11.46} & 3.77           & 3.80           \\
			ORE+TFA\cite{ORE} &14.14 	&28.12 	&2.18 	&0.32 	&6.12 &	4.59 	&4.60 &	15.64 	&28.17 	&3.22 &	0.42 &	6.28 &	4.87 &	4.88 \\
			PROSER+TFA\cite{PROSER}                         & 13.58          & 24.84          & 2.32          & 0.87          & 7.14           & 7.53           & 7.53           & 14.27          & 24.91          & 3.62           & 0.95          & 6.79           & 8.66           & 8.64           \\
			OpenDet+TFA\cite{OpenDet}                        & \textbf{16.01} & \textbf{29.72} & 2.29          & 0.86          & 7.20           & 7.24           & 7.24           & 16.39          & 29.52          & 3.27           & 0.96          & 6.84           & 8.97           & 8.94           \\
			Our FOOD                           & 15.88          & 29.50          & 2.26          & \textbf{2.00} & 5.75           & \textbf{15.76} & \textbf{15.49} & \textbf{16.56} & \textbf{29.55} & 3.58           & \textbf{2.42} & 5.68           & \textbf{18.42} & \textbf{18.02} \\ \midrule[0.8pt]
			\multirow{2}{*}{VOC-COCO}                  & \multicolumn{7}{c|}{3-shot}                                                                                        & \multicolumn{7}{c}{5-shot}                                                                                          \\
			& $mAP_K$        & $mAP_B$        & $mAP_N$       & $AP_U$        & $P_U$          & $R_U$          & $F_U$          & $mAP_K$        & $mAP_B$        & $mAP_N$        & $AP_U$        & $P_U$          & $R_U$          & $F_U$          \\ \midrule[0.8pt]
			TFA\cite{TFA}                                & 16.55          & 28.27          & \textbf{4.81} & 0.00          & 0.00           & 0.00           & 0.00           & 17.13          & 27.71          & 6.56           & 0.00          & 0.00           & 0.00           & 0.00           \\
			DS+TFA\cite{DS}                             & 16.93          & 29.38          & 4.49          & 0.54          & \textbf{12.47} & 3.95           & 3.98           & 17.10          & 27.91          & 6.30           & 0.57          & \textbf{14.15} & 3.86           & 3.89           \\
			ORE+TFA\cite{ORE} &15.74 	&27.41 	&4.23 	&0.52 	&6.49 	&4.91 	&4.92 	&16.21 	&27.32 	&6.29 &0.55 &	6.61 	&4.99 &	5.00 \\
			PROSER+TFA\cite{PROSER}                         & 15.07          & 25.53          & 4.62          & 1.21          & 7.27           & 9.20           & 9.18           & 15.67          & 26.95          & 6.40           & 1.30          & 7.65           & 9.59           & 9.57           \\
			OpenDet+TFA\cite{OpenDet}                        & 16.82          & 29.11          & 4.55          & 1.16          & 7.39           & 9.56           & 9.53           & 17.16          & 27.75          & 6.56           & 1.48          & 7.84           & 11.49          & 11.44          \\
			Our FOOD                           & \textbf{17.56} & \textbf{30.57} & 4.56          & \textbf{2.72} & 6.18           & \textbf{18.93} & \textbf{18.55} & \textbf{18.08} & \textbf{29.47} & \textbf{6.69}  & \textbf{2.92} & 6.06           & \textbf{20.02} & \textbf{19.57} \\ \midrule[0.8pt]
			\multirow{2}{*}{VOC-COCO}                  & \multicolumn{7}{c|}{10-shot}                                                                                       & \multicolumn{7}{c}{30-shot}                                                                                         \\
			& $mAP_K$        & $mAP_B$        & $mAP_N$       & $AP_U$        & $P_U$          & $R_U$          & $F_U$          & $mAP_K$        & $mAP_B$        & $mAP_N$        & $AP_U$        & $P_U$          & $R_U$          & $F_U$          \\ \midrule[0.8pt]
			TFA\cite{TFA}                                & 18.67          & 28.32          & 9.02          & 0.00          & 0.00           & 0.00           & 0.00           & 23.10          & 31.03          & 15.16          & 0.00          & 0.00           & 0.00           & 0.00           \\
			DS+TFA\cite{DS}                             & 19.06          & 28.66          & 9.46          & 0.60          & \textbf{15.41} & 3.75           & 3.78           & 23.40          & 31.53          & \textbf{15.27} & 0.63          & \textbf{15.58} & 3.95           & 3.98           \\
			ORE+TFA\cite{ORE} & 17.98 	&28.01 	&8.75 	&0.59 	&6.77 	&5.13 	&5.14 	&23.07 	&30.47 	&15.17 	&0.66 	&6.89 	&5.51 &	5.52 \\
			PROSER+TFA\cite{PROSER}                         & 17.00          & 25.24          & 8.75          & 1.36          & 7.87           & 10.06          & 10.03          & 21.44          & 28.58          & 14.30          & 1.52          & 7.31           & 12.06          & 11.98          \\
			OpenDet+TFA\cite{OpenDet}                        & 18.53          & 28.36          & 8.70          & 1.82          & 7.38           & 13.89          & 13.77          & 22.93          & 31.61          & 14.02          & 2.64          & 7.52           & 18.07          & 17.82          \\
			Our FOOD                           & \textbf{20.17} & \textbf{30.67} & \textbf{9.48} & \textbf{3.27} & 6.48           & \textbf{21.48} & \textbf{21.00} & \textbf{23.90} & \textbf{33.32} & 14.47          & \textbf{3.80} & 6.68           & \textbf{23.17} & \textbf{22.62} \\ \bottomrule[1pt]
	\end{tabular}}
\end{table*}

\subsubsection{Evaluation Metrics}
The mean average precision (mAP) of known classes ($mAP_K$) is chosen to evaluate the known object detection performance. $mAP_B$ and $mAP_N$ are used to measure the performance for base and novel classes, respectively. To evaluate the unknown detection performance, the average precision ($AP_U$), precision ($P_U$), and recall ($R_U$) are reported. The unknown recall ($R_U$) is a popular metric that is currently concerned by unknown detection \cite{OWDETR}.
%

Precision and recall only focus on one aspect of performance, while F-score can be considered comprehensively for unknown evaluation,
\begin{equation}
	F_U=(1+\beta^2)\frac{P_U\times R_U}{\beta^2\times P_U+R_U},
\end{equation}
where the hyperparameter $\beta$ (the default value is 10 in this paper) indicates the importance of recall relative to precision in the unknown performance evaluation. The calculations of precision and recall rely heavily on labeled data. However, the testing dataset contains many unknown classes $C_U^*\not\in C_U$ that may be detected but not labeled, thus causing evaluation bias. We supply Wilderness Impact (\textbf{WI}) \cite{OpenDet} and Absolute Open-Set Error (\textbf{AOSE}) \cite{OpenDet} to mitigate the evaluation bias of these metrics.
We use WI to measure the degree of unknown objects misclassified to known classes. We also use AOSE to count the number of misclassified unknown objects. Note that the smaller the WI and AOSE values, the better the open-set performance.

\subsubsection{Implementation Details}
Our base detector is Faster R-CNN and ResNet-50 with feature pyramid network (FPN) \cite{FPN} is selected as the backbone. All models are trained using an SGD optimizer with a mini-batch size of 16, a momentum of 0.9, and a weight decay of $1\times 10^{-4}$. The learning rate of 0.02 is used in the base training stage and 0.01 in the few-shot fine-tuning stage. For the CWSC, we set the temperature factor $\alpha=20$ \cite{TFA}, the sparsity probability $\widehat p=0.6$, and the class weight dimension $D=2048$ \cite{TFA}. For the UDL, we use a slope factor $\delta_1=\delta_2=0.09$ and $N_{pos}:N_{neg}=3:12$. In the total loss, we set the trade-off factor $\lambda=1.0$. Note that both the base training stage and the few-shot fine-tuning stage need to optimize the UDL branch. Then we argue that the generalization knowledge learned by the base training stage between the known and unknown classes can be reserved for the few-shot fine-tuning stage. CWSC is only used in the fine-tuning stage.

\subsubsection{Baselines}
We compare our proposed FOOD with the following methods:  OpenDet \cite{OpenDet}, DS \cite{DS}, ORE \cite{ORE}, and PROSER \cite{PROSER} combined with TFA \cite{TFA} for G-FOOD. For dropout sampling (DS) \cite{DS}, the detector is trained with a dropout layer inserted after the second fc layers and we enable it to perform 30 samplings during testing. For PROSER \cite{PROSER}, we remove the proposal’s probability of the ground truth label as the masked probability and match the masked probability with the unknown class $C_U=\{K+1\}$, forcing the dummy unknown classiﬁer to output the probability for unknown rejection.
For ORE \cite{ORE} and OpenDet \cite{OpenDet}, we use the official codes combined with a two-stage fine-tuning strategy (TFA \cite{TFA}) for generalized few-shot open-set detection.
We also present the FSOD results of TFA as a baseline to determine whether optimizing the dummy unknown class will reduce the performance of the known classes.
Moreover, all methods employ the same ResNet-50 with FPN as the backbone for a fair comparison and we report the average results of 10 random runs for all comparison methods.  

\begin{table*}[]
	\caption{The Open-set Performance (WI and AOSE) of Different Methods on VOC10-5-5 and VOC-COCO Dataset Settings. $\downarrow$ Indicates That the Smaller the Evaluation Metrics, the Better the Performance. \textbf{Bold} Numbers denote Superior Results. For a Fair Comparison, We Report the Average Results of 10 Random Runs for All Comparison Methods.}\label{table_01}
	\renewcommand\arraystretch{0.95}\centering
		\begin{tabular}{l|cccccc|cccccc}
			\toprule[1pt]
			\textbf{VOC10-5-5}   & \multicolumn{6}{c|}{$WI\downarrow$}                                                                     & \multicolumn{6}{c}{$AOSE\downarrow$}                                                                                  \\ \midrule[0.8pt]
			Method      & 1             & 3             & 5             & 10            & 30            & Mean          & 1               & 3               & 5               & 10               & 30               & Mean            \\ \midrule[0.8pt]
			TFA\cite{TFA}         & 10.69         & 10.13         & 9.99          & 9.87          & 8.93          & 9.92          & 1308.40         & 1335.40         & 1256.10         & 1267.20          & 1237.00          & 1280.82         \\
			DS+TFA\cite{DS}      & 9.14          & 9.08          & 8.97          & 8.81          & 8.55          & 8.91          & 772.60          & 969.90          & 990.60          & 1025.70          & 1138.50          & 979.46          \\
			ORE+TFA\cite{ORE}     & 9.54          & 9.88          & 10.16         & 9.65          & 9.64          & 9.77          & 930.30          & 1058.70         & 1019.70         & 1063.70          & 1159.10          & 1046.30         \\
			PROSER+TFA\cite{PROSER}  & 11.15         & 10.45         & 10.65         & 10.29         & 9.62          & 10.43         & 994.60          & 1021.70         & 1009.80         & 956.70           & 996.40           & 995.84          \\
			OpenDet+TFA\cite{OpenDet} & 10.47         & 9.27          & 9.01          & 8.50          & 7.54          & 8.96          & 867.30          & 954.50          & 1031.50         & 1021.40          & 1103.50          & 995.64          \\
			Our FOOD    & \textbf{6.96} & \textbf{7.83} & \textbf{7.59} & \textbf{6.99} & \textbf{6.57} & \textbf{7.19} & \textbf{598.60} & \textbf{859.00} & \textbf{908.00} & \textbf{900.20}  & \textbf{923.90} & \textbf{837.94} \\ \midrule[0.8pt]
			\textbf{VOC-COCO}    & \multicolumn{6}{c|}{$WI$}                                                                     & \multicolumn{6}{c}{$AOSE$}                                                                                  \\ \midrule[0.8pt]
			Method      & 1             & 3             & 5             & 10            & 30            & Mean          & 1               & 3               & 5               & 10               & 30               & Mean            \\ \midrule[0.8pt]
			TFA\cite{TFA}         & 10.73         & 10.77         & 11.36         & 11.40         & 10.48         & 10.95         & 1441.80         & 1566.70         & 1673.30         & 1732.20          & 2294.10          & 1741.62         \\
			DS+TFA\cite{DS}      & 9.15          & 9.56          & 9.91          & 10.13         & 9.84          & 9.72          & 711.60          & 1007.60         & 1110.10         & 1336.40          & 1892.90          & 1211.72         \\
			ORE+TFA\cite{ORE}     & 12.08         & 11.53         & 12.30         & 11.65         & 11.22         & 11.76         & 1087.00         & 1226.00         & 1344.00         & 1463.20          & 1867.00          & 1397.44         \\
			PROSER+TFA\cite{PROSER}  & 11.68         & 12.46         & 12.56         & 12.47         & 12.00         & 12.23         & 925.30          & 1165.10         & 1165.90         & 1160.00          & 1561.60          & 1195.58         \\
			OpenDet+TFA\cite{OpenDet} & 9.82          & 9.92          & 9.55          & 9.83          & 9.02          & 9.63          & 690.90          & 1053.00         & 1176.90         & 1400.60          & 1818.00          & 1227.88         \\
			Our FOOD    & \textbf{6.78} & \textbf{6.95} & \textbf{7.37} & \textbf{7.95} & \textbf{8.13} & \textbf{7.44} & \textbf{485.00} & \textbf{700.40} & \textbf{859.00} & \textbf{1099.30} & \textbf{1480.90} & \textbf{924.92} \\ \bottomrule[1pt]
	\end{tabular}
\end{table*}

\subsection{Results}
\subsubsection{\textbf{VOC10-5-5}} In Table \ref{table_0}, we compare our FOOD with several OSOD methods combined with TFA on VOC10-5-5 dataset settings. Our FOOD outperforms other methods across all shots on the unknown metrics $AP_U$, $R_U$, and $F_U$. We achieve 0.65$\sim$1.27 point improvement in $AP_U$ over the best comparison method, around 5.89$\sim$13.99 point improvement in $R_U$, and around 3.38$\sim$10.63 point improvement in $F_U$. Simultaneously, our method outperforms other methods on the $mAP_N$ of novel classes, which demonstrates its effectiveness for G-FOOD. The unknown F-score of our FOOD achieves a significant improvement over the second best, which verifies that our method alleviates the over-fitting issue and evidently improves the model's generalization ability for unknown rejection in few-shot scenes.  
However, in VOC10-5-5 settings, the detector may have seen unknown objects and treated them as background during training, which causes the evaluation bias for G-FOOD. To balance the above bias, we conduct experiments on VOC-COCO dataset settings, where the unknown classes are barely seen in the training data.

\subsubsection{\textbf{VOC-COCO}} In Table \ref{table_1}, we carry out experiments on VOC-COCO dataset settings. Only looking at the rejection performance of the unknown class ($AP_U$, $P_U$, $R_U$, and $F_U$), our method presents several significant advantages, especially in unknown recall $R_U$ and F-score $F_U$. For example, the highest F-score 22.62\% of the unknown class is obtained by our FOOD,  which illustrates that our method has a better unknown object rejection ability than other methods. Simultaneously, in extremely low shots (1, 2, 3, 5-shot) where the over-fitting problem is easier to occur than in high shots (10, 30-shot), our method achieves 7.96\%, 9.08\%, 9.02\%, and 8.13\% unknown F-score $F_U$ improvements than the second best method respectively, which demonstrates the effectiveness of our method in suppressing the over-fitting issue caused by few training samples of known classes. 

Compared with the baseline TFA on novel classes, the $mAP_N$ of our method is similar to TFA. Meanwhile, the $mAP_K$ and $mAP_B$ of our method are higher than TFA, which verifies that our method not only improves the open-set detection performance of the unknown class, but also slightly improves the performance of known classes $C_K=C_B\cup C_N$ for the closed-set few-shot evaluation. Look at other methods, although
DS+TFA, PROSER+TFA, and OpenDet+TFA achieve comparable closed-set metrics ($mAP_K$, $mAP_B$, and $mAP_N$), the open-set performance is poor. Overall, the proposed FOOD surpasses other methods and thus is of merit. The class weight sparsification and the unknown decoupling training effectively expand model's generalization capability for open-set detection in few-shot conditions, which brings performance improvement.

Our FOOD method often achieves greater unknown recall at a minor cost of unknown precision. With the precision and recall for unknown evaluation, it appears that these are calculated from the classes that are labeled in the respective datasets but treated as unknown classes. However, there may still be unknown objects that are not labeled in these datasets. If the method detects these, it may lead to lower precision as it will be considered a false-positive (FP) unknown. This is an inherent weakness in using these metrics with unknown classes. We report several new evaluation metrics including $WI$ and $AOSE$ to evaluate the open-set performance of our method, which can mitigate the evaluation bias caused by AP, precision, and recall. As illustrated in Table \ref{table_01}, we can see that our method achieves the best mean performance of WI (7.19 and 7.44) and AOSE (837.94 and 924.92) in VOC10-5-5 and VOC-COCO dataset settings, respectively, which demonstrates that our method has a strong ability for unknown rejection. Thus, the effectiveness of our method is further verified.

\subsubsection{\textbf{LVIS315-454-461}} LVIS dataset has the characteristics of numerous classes (1230 classes) and long-tail distribution, which is very challenging. We treat the frequent classes as
base known classes, the rare classes as novel known classes, and the common classes as unknown classes. As shown in Table \ref{table_001}, weight sparsification helps the model suppress overfitting well, so the few-shot closed-set detection can maintain good experimental results ($mAP_B$ and $mAP_N$). At the same time, compared with the second-best method, our model achieves good open-set performance improvements (0.96\%, 3.88\%, 4.03, and 343 in $AP_U$, $F_U$, $WI$, and $AOSE$, respectively). This proves that the decoupling optimization improves the model's generalization ability for unknown rejection in few-shot scenes, which is meaningful for real-world long-tail distribution applications.

\subsection{Ablation Studies}
We conduct comprehensive ablation studies on the 10-shot VOC-COCO dataset setting.
\subsubsection{Effectiveness of different modules}
We perform the ablation studies on different modules (CWSC and UDL) in Table \ref{table_2}, where the classical FSOD framework TFA \cite{TFA} is used as the baseline. It can be seen that CWSC boosts the detection performance of the base classes ($AP_B$) and the novel classes ($AP_N$) simultaneously, which demonstrates the effectiveness of CWSC in alleviating the over-fitting problem. When exploring the influence of UDL, we find that the branch of UDL improves the detection ability for the closed-set known classes in the 10-shot setting. Simultaneously, UDL is a necessary condition for the unknown rejection of our FOOD. The best unknown detection result $AP_U=3.27\%$ is obtained by the cooperation of two modules, although the detection performance of known classes is slightly degraded compared to the UDL alone ($\downarrow0.10\%$), this is acceptable in terms of the overall results. Compared with the third line (the linear classifier with a dropout layer+UDL) and the fourth line (CWSC+UDL), the unknown metrics including F-score, $WI$, and $AOSE$ improve by 4.56\%, 0.63, and 140.80, respectively, which demonstrates that the unknown rejection performance benefits from our CWSC. The class weight sparsification significantly enhances the model's generalization ability for unknown rejection in few-shot scenes.

\begin{table}[]
	\caption{The generalized few-shot Open-set Performance on LVIS315-454-461 Dataset Settings.}\label{table_001}
	\resizebox{\linewidth}{!}{
	\begin{tabular}{l|cccccc}
		\toprule[1pt]
		\textbf{LVIS315-454-461} & $mAP_B$$\uparrow$ & $mAP_N$$\uparrow$ & $AP_U$$\uparrow$ & $F_U$$\uparrow$              & $WI$$\downarrow$          & $AOSE$$\downarrow$ \\ \midrule[0.8pt]
		TFA\cite{TFA}             & 23.08   & 12.24   & 0      & 0     & 16.78 & 2261.20 \\
		DS+TFA\cite{DS}          & 23.19   & 12.79   & 0.28   & 3.87  & 15.45 & 2107.50 \\
		ORE+TFA\cite{ORE}         & 22.41   & 12.10   & 0.26   & 3.91  & 15.21 & 2217.30 \\
		PROSER+TFA\cite{PROSER}      & 23.07   & 12.48   & 0.57   & 6.27  & 17.75 & 2158.80 \\
		OpenDet+TFA\cite{OpenDet}     & 23.14   & 12.43   & 0.62   & 6.58  & 14.83 & 1923.50 \\
		Our FOOD        & \textbf{24.51}   & \textbf{13.66}   & \textbf{1.58}   & \textbf{10.46} & \textbf{10.80}  & \textbf{1580.50} \\ \bottomrule[1pt]
	\end{tabular}}
\end{table}


\begin{table}[]
	\centering
	\caption{The Ablation Study of Different Modules (Average of 10 Random Runs). \textbf{Bold} Numbers denote Superior Results.}	\label{table_2}
	\resizebox{\columnwidth}{!}{
	\begin{tabular}{cc|cccccc}
		\toprule[1pt]
		CWSC              & UDL             & $mAP_B$$\uparrow$        & $mAP_N$$\uparrow$       & $AP_U$$\uparrow$        & $F_U$$\uparrow$          & $WI$$\downarrow$          & $AOSE$$\downarrow$          \\ \midrule[0.8pt]
		\multicolumn{2}{l|}{Baseline (TFA)} & 28.32          & 9.02          & 0             & 0              & 11.40          & 1732.20         \\
		\ding{51}                 &                 & 28.97          & 9.44          & 0             & 0              & 8.97          & 1658.90         \\
		& \ding{51}               & \textbf{30.96} & \textbf{9.58} & 2.35          & 16.44          & 8.58          & 1240.10         \\
		\ding{51}                 & \ding{51}               & 30.67          & 9.48          & \textbf{3.27} & \textbf{21.00} & \textbf{7.95} & \textbf{1099.30} \\ \bottomrule[1pt]
	\end{tabular}}
\end{table}

\begin{table}[]
	\centering
	\caption{The Ablation Study of Different Fine-Tuning Methods (Average of 10 Random Runs). \textbf{Bold} Numbers denote Superior Results.}	\label{table_3}
	\resizebox{\columnwidth}{!}{
	\begin{tabular}{l|cccccc}
		\toprule[1pt]
		Fine-tuning methods & $mAP_B$$\uparrow$        & $mAP_N$$\uparrow$        & $AP_U$$\uparrow$        & $F_U$$\uparrow$          & $WI$$\downarrow$          & $AOSE$$\downarrow$          \\ \midrule[0.8pt]
		LP                  & 30.55          & 9.22           & 3.14          & 20.47          & 8.67          & 1188.10          \\
		FT                  & 20.59          & \textbf{10.03} & 0.78          & 5.19           & 8.30           & 1192.00          \\
		FT+GDL              & 30.42          & 9.78           & 3.20           & 20.83          & 8.12          & 1125.20          \\
		LP-FT               & 30.35          & 9.35           & 3.12          & 20.45          & 8.59          & 1203.40          \\
		LP-FT+GDL           & \textbf{30.67} & 9.48           & \textbf{3.27} & \textbf{21.00} & \textbf{7.95} & \textbf{1099.30} \\ \bottomrule[1pt]
	\end{tabular}}
\end{table}

\subsubsection{Effectiveness of different fine-tuning methods}
In Table \ref{table_3}, we carefully evaluate several fine-tuning methods to pick the most appropriate way. Linear-probing (LP) and fine-tuning (FT) \cite{ICLR2022oral} have been widely used in transfer learning to alleviate the over-fitting problem.
Gradient decoupled layer (GDL) \cite{DeFRCN} is an auxiliary fine-tuning strategy, which conducts a stop gradient for RPN and a scaled gradient (scale=0.001 in this paper) for RCNN. As a hard-freezing method, LP can preserve the general knowledge of the base training stage, thus it achieves a competitive detection result in Table \ref{table_3}. However, FT fine-tunes the entire model to fit the few-shot closed-set training data. The model gradually forgets the base classes and destroys the general knowledge during the few-shot fine-tuning stage. Thus, we can see that compared FT with LP, the results of the base classes degrade by 9.96\%, and the unknown class descends by 2.36\%. The advantage of FT is the better performance of the novel classes $mAP_N$ than other approaches. LP-FT exploits the advantages of LP and FT, thus it achieves a compromised G-FOOD result. 

Compared with FT, FT+GDL achieves better results in $mAP_K$ and $AP_U$. 
This indicates that GDL reserves the general knowledge by the gradient decoupling training, which uses a scaled gradient to slowly update the parameters of the backbone network. We view the GDL as a soft-freezing method, which enables the backbone to slowly fit the closed-set data while retaining the ability to extract the generalization features. GDL improves the performance of the closed-set object detection and achieves a competitive unknown rejection performance.
Based on the above analysis, we adopt a \textbf{hard-soft} combination approach (LP-FT+GDL), which first trains the last linear layers of the model while freezing other parameters. And then we fine-tune the model in a soft-freezing way. As illustrated in Table \ref{table_3}, LP-FT+GDL outperforms other approaches in terms of $mAP_B$, $AP_U$, $F_U$, $WI$, and $AOSE$ which verifies its effectiveness.


\begin{table}[]
		\centering
	\caption{The Ablation Study on Different Sampling Methods (Average of 10 Random Runs). \textbf{Bold} Numbers denote Superior Results.}	\label{table_4}
	\resizebox{\columnwidth}{!}{
	\begin{tabular}{l|cccccc}
		\toprule[1pt]
		Simpling methods          & $mAP_B$$\uparrow$        & $mAP_N$$\uparrow$       & $AP_U$$\uparrow$        & $F_U$$\uparrow$          & $WI$$\downarrow$          & $AOSE$$\downarrow$           \\ \midrule[0.8pt]
		$Min max-probability$     & 30.30          & 9.30          & 3.05          & 20.34          & 10.08         & 1332.50          \\
		$Min(l_{C_U})$            & 29.39          & \textbf{9.66} & 2.73          & 17.29          & 10.16         & 1448.50          \\
		$Max(entropy)$            & 29.17          & 9.41          & 2.79          & 18.18          &  9.24         & 1573.00          \\
		$Max(E(x,b)$)             & 30.57          & 9.31          & 3.08          & 20.43          & 8.10          & 1288.80          \\
		$Max(E(x,b)_{c_i\neq u}$) & \textbf{30.67} & 9.48          & \textbf{3.27} & \textbf{21.00} & \textbf{7.95} & \textbf{1099.30} \\ \bottomrule[1pt]
	\end{tabular}}
\end{table}

\subsubsection{Effectiveness of different sampling methods} 
The sampling rules for the positive pseudo-unknown samples to train the UDL branch are as follows:
\begin{itemize} 
	\item $Min\;max$-$probability$ \cite{OpenDet}: the samples are sorted in ascending order by the maximum predicted value across all classes, and top-$k$ samples are chosen.
	\item $Min(l_{C_U})$:  the samples are sorted in ascending order by the unknown logit, and top-$k$ samples are chosen.
	\item $Max(entropy)$: the samples are sorted in descending order by the entropy score, and top-$k$ samples are chosen.
	\item $Max(E(x,b)_{c_i\neq C_U})$: the samples are sorted in descending order by the conditional energy score. We select top-$k$ samples for optimization.
\end{itemize}

As presented in Table \ref{table_4}, the conditional energy-based sampling method $Max(E(x,b)_{c_i\neq C_U})$ outperforms other methods, which demonstrates its effectiveness for the pseudo-unknown sample selection. Furthermore, the absence of real unknown training samples causes the unknown class $C_U$ to become a distractor, thus the results of our $Max(E(x,b)_{c_i\neq C_U})$ are better than $Max(E(x,b))$. Simultaneously, we can see that these energy-based sampling methods ($Max(E(x,b))$ and $Max(E(x,b)_{c_i\neq C_U})$) perform better than other sampling methods for open-set detection ($WI$ and $AOSE$), which proves that the energy score is an excellent uncertainty metric for the pseudo-unknown sample selection in the optimization of our UDL branch.

\begin{figure}[!t]
	\centering
	\includegraphics[width=8.8cm]{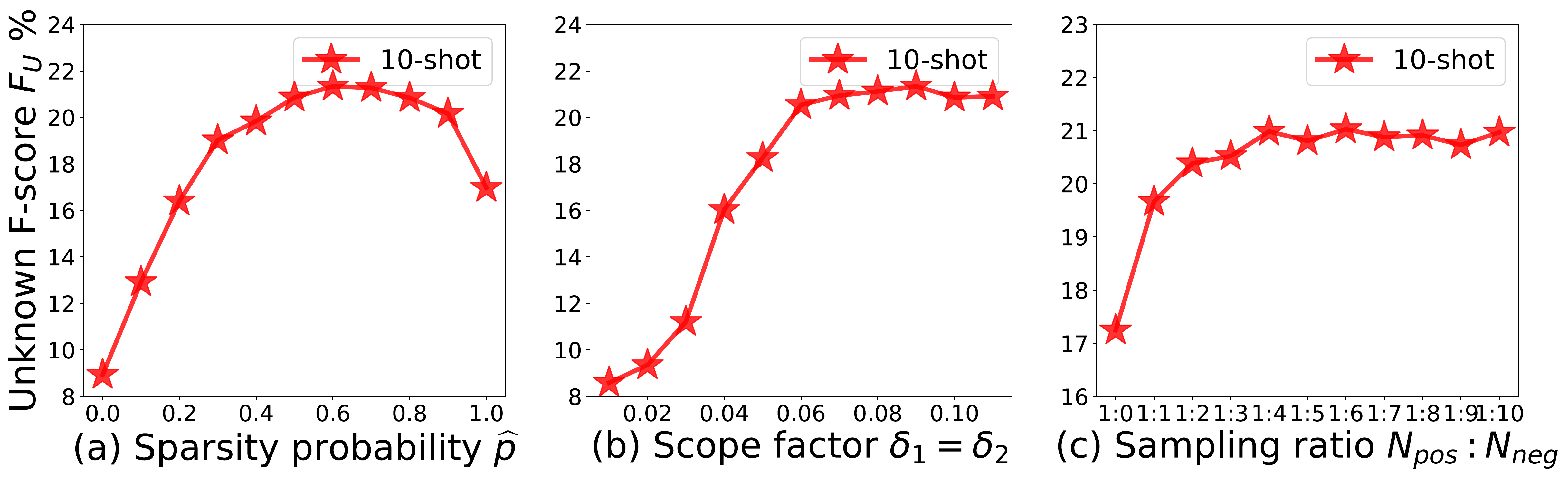}
	\caption{Effect of different sparsity probabilities $\widehat p$, scope factors $\delta_1=\delta_2$ and sampling ratios $N_{pos}:N_{neg}$ on the 10-shot VOC-COCO dataset setting.}\label{fig4}
\end{figure}

\begin{table}[]
		\centering
	\caption{The Ablation Study of Different (a) Backbones and (b) FSOD Methods (Average of 10 Random Runs). \textbf{Bold} Numbers denote Superior Results.}	\label{table_5}
	\resizebox{\columnwidth}{!}{
	\begin{tabular}{l|cccccc}
		\toprule[1pt]
		\textbf{(a)} Backbones  & $mAP_B$$\uparrow$        & $mAP_N$$\uparrow$       & $AP_U$$\uparrow$        & $F_U$$\uparrow$              & $WI$$\downarrow$          & $AOSE$$\downarrow$          \\ \midrule[0.8pt]
		ResNet-50  & 30.67          & 9.48          & 3.27          & 21.00          & 7.95 & 1099.30 \\
		ResNet-101 & 31.48          & 10.00            & 3.32          & 21.78           & 7.56          & 1035.00            \\
		Swin-T\cite{Swin-T}     & \textbf{40.49} & \textbf{10.80} & \textbf{3.41} & \textbf{22.32} & \textbf{7.27}          & \textbf{867.00}            \\ \midrule[0.8pt]
		\textbf{(b)} FSOD methods	\\ \midrule[0.8pt]
		TFA-based FOOD \cite{TFA}    & 30.67   & 9.48    & 3.27   & 21.00 & 7.95 & 1099.30 \\ 
		DeFRCN-based FOOD \cite{DeFRCN} & 30.91   & 13.82   & 7.03   & 27.14 & 6.47 & 918.00  \\
		FSRDD-based FOOD \cite{FSRDD}  & \textbf{31.05}   & \textbf{15.47}   & \textbf{7.12}   & \textbf{28.94} & \textbf{6.25} & \textbf{909.90}   \\ \bottomrule[1pt]
	\end{tabular}}
\end{table}

\begin{figure}[t]
	\centering
	\includegraphics[width=8cm]{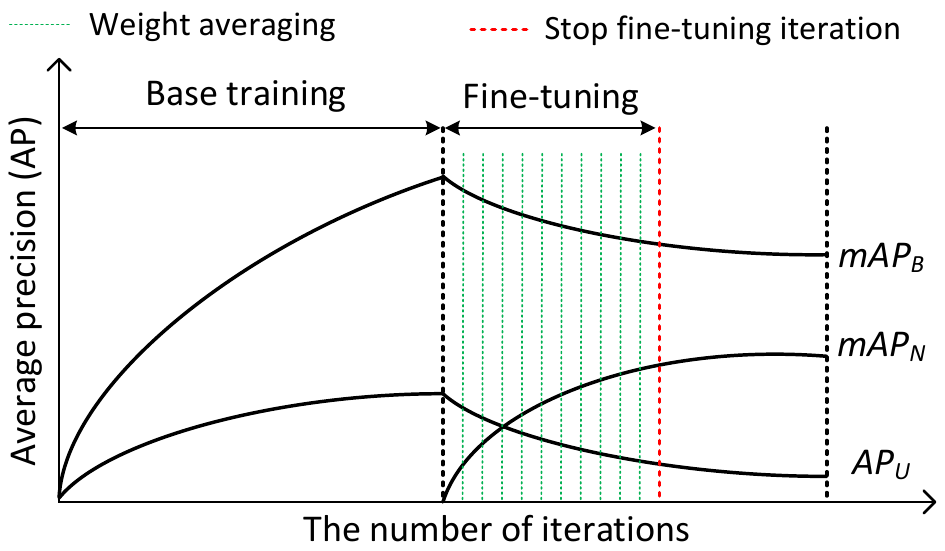}
	\caption{The relationship between iteration and performance metrics ($mAP_B$, $mAP_N$, and $AP_U$) for our G-FOOD method. We are hard to select a proper stop fine-tuning iteration to balance the performance of base classes, novel classes, and the unknown class.}
	\label{fig6}
\end{figure}

\subsubsection{Sparsity probability, scope factor, and sampling ratio} 
Fig. \ref{fig4} presents the visualization of different sparsity probabilities, scope factors, and sampling ratios on the 10-shot VOC-COCO dataset setting. We fix the ratio of positive and negative samples $1:4$ to explore the performance of different sparsity probabilities. As presented in Fig. \ref{fig4}(a), when the sparsity probability is set to 0.6, the unknown F-score $F_U$ arrives at 21.34\%, which is higher than other settings. Subsequently, we fix $\widehat p=0.6$ to iterate over different scope factors and sampling ratios, respectively. As shown in Fig. \ref{fig4} (b), when the scope factor is set to 0.09, the unknown F-score is best. Therefore, we set $\delta_1=\delta_2=0.09$. As we can see from Fig. \ref{fig4}(c), starting from a sampling ratio of $1:4$, the value of unknown F-score begins to be stabilized. The sampling ratio of $1:6$ seems to perform the best but our usual default value of $1:4$ is close to the optimal. Moreover, if the pseudo-unknown samples selected from the background do not participate in the optimization of the UDL branch ($1:0$), the unknown rejection performance drops significantly.


\subsubsection{Effectiveness of different backbones} 
We use ResNet-101 \cite{Resnet101} and swin transformer (Swin-T) \cite{Swin-T} as the backbones in Table \ref{table_5} (a), and then compare them with ResNet-50. ResNet-101 tends to perform better than ResNet-50, which presents that our method benefits from the deeper backbone. While employing a more powerful transformer-based backbone (Swin-T), our proposed method achieves additional improvements.

\begin{figure*}[!t]
	\centering
	\includegraphics[width=16.5cm]{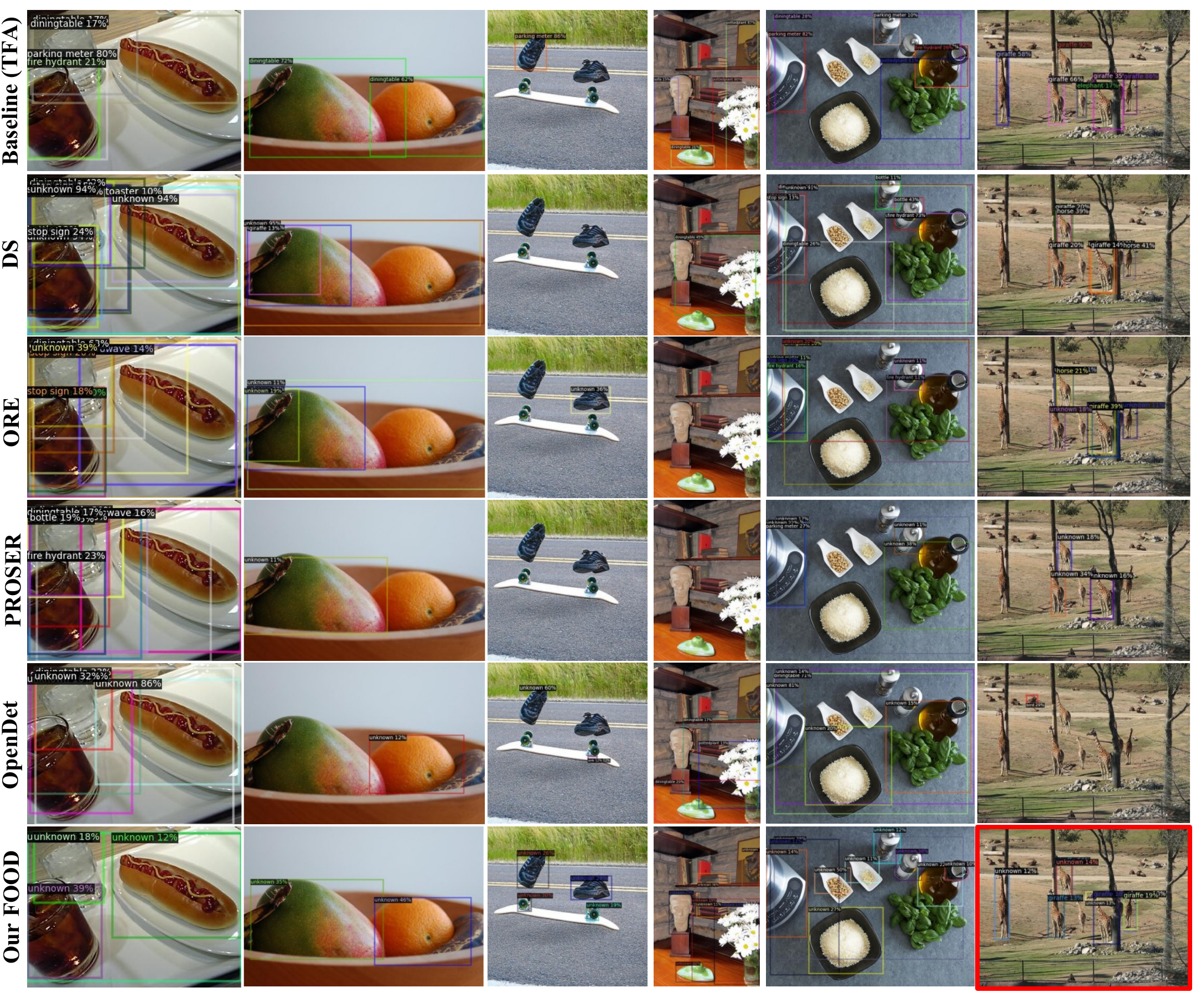}
	\caption{The visualization results (10-shot VOC-COCO setting). We visualize the bounding boxes with a score larger
		than 0.1. Our FOOD can detect more unknown objects than other methods. Red box is the failure case, several giraffes (novel class) are misidentified as the unknown class.}\label{fig5}
\end{figure*}

\subsubsection{Does the unknown-class placeholder hurt the accuracy of the few-shot object detection?} \textbf{No}.
As illustrated in the first and third lines of Table \ref{table_2}, when the UDL branch introduces the unknown-class placeholder into the baseline TFA, the performance of novel classes has achieved 0.56\% improvement, which proves that the dummy placeholder of the unknown class does not hurt the detection performance for few-shot novel classes. Furthermore, UDL and CWSC can feasibly be applied to most few-shot object detectors such as TFA \cite{TFA}, DeFRCN \cite{DeFRCN}, and FSRDD \cite{FSRDD}. As illustrated in Table \ref{table_5} (b), when incorporating UDL and CWSC into the recent FSRDD, few-shot performance and open-set performance achieve evidently improvements compared to the TFA-based few-shot open-set detection method, which demonstrates that our method is agnostic to the underlying few-shot object detection (FSOD) methods. It benefits from the high performance obtained by FSOD methods.


\begin{table}[]
	\caption{The Performance of Weight Averaging (WA) (Average of 10 Random Runs). \textbf{Bold} Numbers denote Superior Results.}	\label{table_6}
	\resizebox{\columnwidth}{!}{
	\begin{tabular}{l|cccccc}
		\toprule[1pt]
		VOC-COCO    & $mAP_B$$\uparrow$        & $mAP_N$$\uparrow$       & $AP_U$$\uparrow$        & $F_U$$\uparrow$              & $WI$$\downarrow$          & $AOSE$$\downarrow$          \\ \midrule[0.8pt]
		Our FOOD    & 30.67          & \textbf{9.48} & 3.27          & 21.00          & 7.95          & 1099.30         \\
		Our FOOD+WA & \textbf{37.89} & 8.76          & \textbf{4.13} & \textbf{24.23} & \textbf{7.15} & \textbf{856.30} \\ \bottomrule[1pt]
	\end{tabular}}
\end{table}

\subsubsection{Weight averaging}
As shown in Fig. \ref{fig6}, the performance trends of the base classes $AP_B$ and the unknown classes $AP_U$ conflict with the novel classes $AP_N$ in the fine-tuning stage of our FOOD method. Therefore, it is difficult for us to choose a suitable stop fine-tuning iteration that can make the base, novel, and unknown classes all perform best. Motivated by weight averaging (WA) \cite{WA} that is a simple ensemble method, but it achieves the state-of-the-art performance in domain generalization \cite{ERM, DWA, DiWA}. We present WA to approximate the optimal model. The WA is defined as:  
\begin{equation}\label{eq11}
	\begin{split}
	\theta_{WA}(\mathcal{L}(D_{tr}\cup \mathcal{S}_{pos}\cup \mathcal{S}_{neg}))=\;\;\;\;\;\;\;\;\;\;\;\;\;\;\;\;\;\;\;\;\;\\ \frac1{H+1}({\textstyle\sum_{h=1}^H}\theta_h(\mathcal{L}(D_{tr}\cup \mathcal{S}_{pos}\cup \mathcal{S}_{neg}))+\theta_{final\;model}),
	\end{split}
\end{equation}
where ${\{\theta_h\}}_{h=1}^H$ represents the weights of equidistant dense sampling in a single run and $\theta_{final\;model}$ denotes the final output weights in the above single run.  WA uses the idea of ensemble learning to balance the representation bias between the base, novel, and unknown classes. 
As illustrated in Table \ref{table_6}, when incorporating with WA (the model sampling step is 100 iterations), the evaluation metrics ($mAP_K$, $mAP_B$, $AP_U$, $F_U$, $WI$, and $AOSE$) of our FOOD achieve evidently improvements, which demonstrates its effectiveness. However, the drawback is that WA slightly decreases the performance of the novel classes $mAP_N$. The main reason is that WA hurts the novel-class performance through the poor weight integration in low fine-tuning iterations. However, if you'd like to significantly improve the detection performance of the base classes and the unknown class at the expense of a little performance for novel classes, WA is a good choice.

\subsection{Visualization}
We provide qualitative visualizations of
the detected unknown objects on the 10-shot VOC-COCO dataset setting in
Fig. \ref{fig5}. We can observe that other methods easily recognize unknown objects as known classes and cannot detect unknown objects frequently. Compared with other methods, our FOOD can achieve better unknown rejection performance, which demonstrates that our method benefits from the class weight sparsification and the unknown decoupling training. However, our method easily identifies the few-shot known objects as the unknown class. As shown in the red box of Fig. \ref{fig5}, several giraffes (novel class) are misidentified as the unknown class. Simultaneously, this phenomenon can also be seen from the quantitative analysis that our method shows slightly low unknown precision $P_U$, as illustrated in Table \ref{table_1}. This situation is a limitation of the dummy class-based method. We argue that introducing Intersection over Union (IoU) constraints for unknown optimization may be a good solution, which will be further researched in the future.

\section{Conclusion}
In this paper, we propose a new task named generalized few-shot open-set object detection (G-FOOD) and build the first benchmark. To tackle the challenging G-FOOD task, we propose a simple method incorporating several tricks in Faster R-CNN with two novel modules: class weight sparsification classifier (CWSC) and unknown decoupling learner (UDL). The CWSC is developed to decrease the co-adaptability between the class and its neighboring classes during the training process, thereby improving the model's generalization ability for open-set detection in few-shot scenes. Alongside, the UDL branch is employed to detect unknown objects in few-shot scenes without depending on the class prototype, threshold, and pseudo-unknown sample generation. Compared with other OSOD methods in few-shot scenes, our method achieves state-of-the-art results on different shot settings of VOC10-5-5, VOC-COCO, and LVIS315-454-461 datasets. In the future, we'd like to study the prompt learning \cite{PL} to solve the challenging G-FOOD task.

{\appendix \label{appendix}
We present the derivation of Eq. (\ref{eq1})-(\ref{eq4}). In a linear regression task, assume $y\in\mathbb{R}^{N}$ is the ground truth label, the model tries to find a $w\in\mathbb{R}^{D}$ to minimize 
\begin{equation}\label{eq1_new}
	\vert\vert y-\alpha \frac X{\vert\vert X\vert\vert}\frac w{\vert\vert w\vert\vert}\vert\vert^{2}.
\end{equation}
We set $\overset-x=X/\vert\vert X\vert\vert$ and $\overset-w=w/\vert\vert w\vert\vert$. When the weight sparsification is adopted, the objective function becomes 
\begin{equation}\label{eq2_new}
	\begin{split}
		&\underset w{minimize}\;{\mathbb{E}}_{R\sim\mathrm{Bernoulli}(p)}\lbrack{\vert\vert y-\alpha \overset-x(R\ast\overset-w)\vert\vert}^2\rbrack\\
		&=\underset w{minimize}{\vert\vert y-\alpha p\overset-x\overset-w\vert\vert}^2+\alpha^2(1-p)^2{\vert\vert \tau\overset-w\vert\vert}^2\\
		&=\underset w{minimize}{\vert\vert y-\alpha p\overset-x\overset-w\vert\vert}^2+\alpha^2(p^2-2p){\vert\vert \tau\overset-w\vert\vert}^2+\alpha^2{\vert\vert \tau\overset-w\vert\vert}^2,
	\end{split}
\end{equation}
where $\tau=({diag(\overset-x^{\mathrm T}\overset-x)}{)}^{1/2}$. We view the above formula as a function of $p$. Then, this can reduce to
\begin{equation}\label{eq3_new}
	\underset w{minimize}{\vert\vert y-\alpha p\overset-x\overset-w\vert\vert}^2+\alpha^2(p^2-2p){\vert\vert \tau\overset-w\vert\vert}^2.
\end{equation}
This can reduce to
\begin{equation}\label{eq4_new}
	\underset w{minimize}{\vert\vert y-\alpha p\overset-x\overset-w\vert\vert}^2-\alpha^2p(2-p){\vert\vert \tau\overset-w\vert\vert}^2.
\end{equation}
This can reduce to
\begin{equation}\label{eq5_new}
	\underset w{minimize}{\vert\vert y-\alpha p\overset-x\overset-w\vert\vert}^2-4\alpha^2\frac p2(1-\frac p2){\vert\vert \tau\overset-w\vert\vert}^2.
\end{equation}
This can reduce to 
\begin{equation}\label{eq6_new}
	\underset w{minimize}{\vert\vert y-\alpha p\overset-x\overset-w\vert\vert}^2+\alpha^2p(1-p){\vert\vert\tau\overset-w\vert\vert}^2. 
\end{equation}
We set $\widetilde w=\alpha p\overset-w$, then 
\begin{equation}\label{eq7_new}
	\underset w{minimize}{\vert\vert y-\overset-x\widetilde w\vert\vert}^2+\underbrace{\frac{(1-p)}p\vert\vert\tau\widetilde\omega\vert\vert^2}_{\tiny Regularization\;term}.
\end{equation}}
\bibliographystyle{IEEEtran}
\bibliography{ieeetrans.bib}


\begin{IEEEbiography}[{\includegraphics[width=1in,height=1.25in,clip,keepaspectratio]{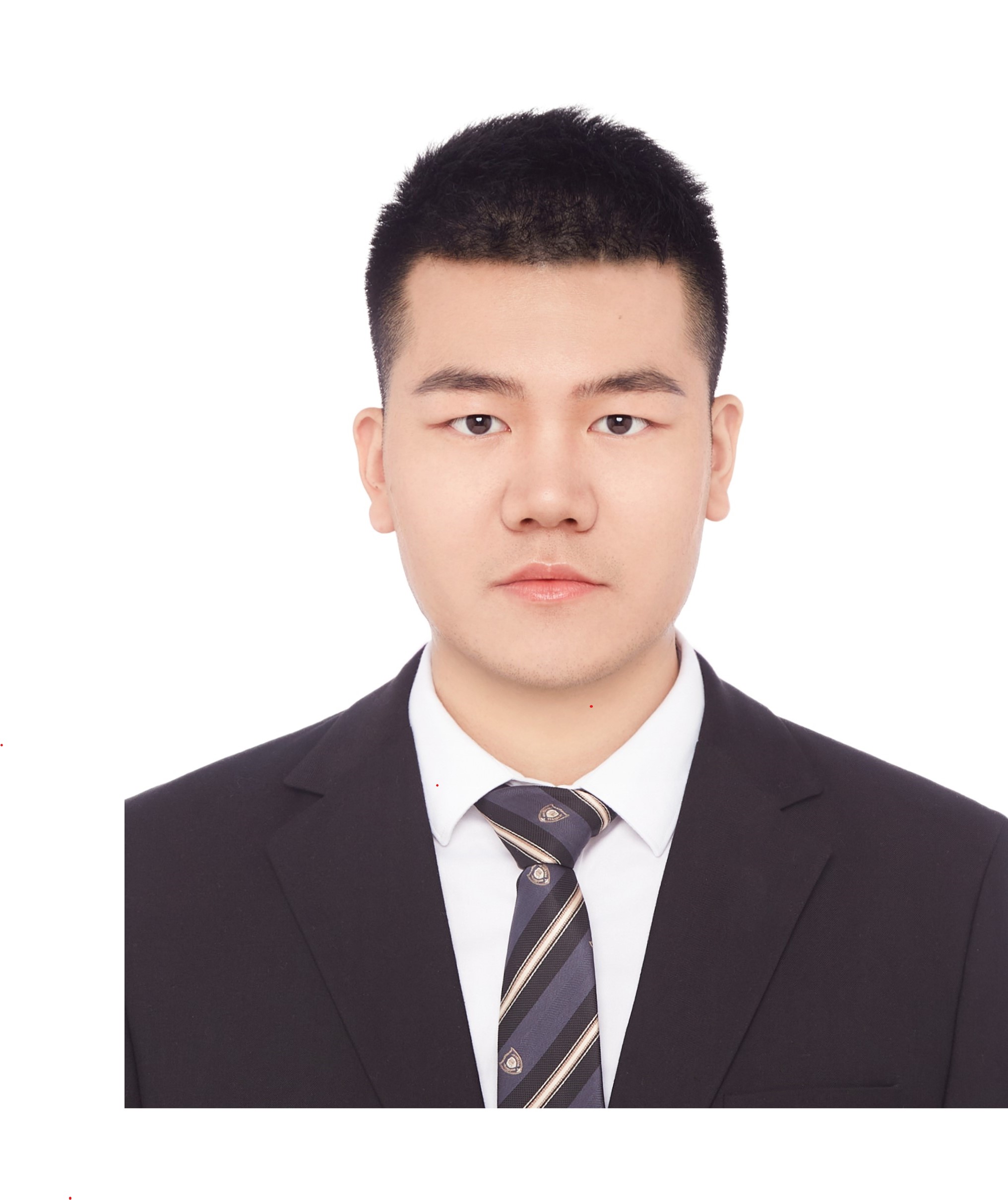}}]
	{Binyi Su} received the B.S. degree in intelligent 
	science and technology from the Hebei
	University of Technology, Tianjin, China, in 2017, and the M.S degree in control engineering from the Hebei
	University of Technology, Tianjin, China, in 2020.
	
	He is currently pursuing the Ph.D. degree in computer science and technology from Beihang University, Beijing, China.
	His current research interests include computer vision and pattern recognition.
\end{IEEEbiography}
%
\begin{IEEEbiography}[{\includegraphics[width=1in,height=1.25in,clip,keepaspectratio]{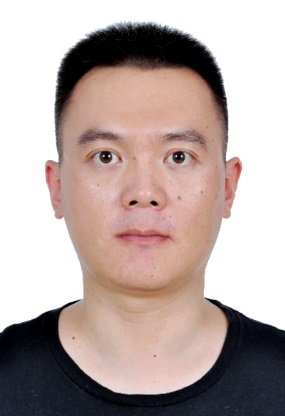}}]{Hua Zhang} received the Ph.D. degrees in computer science from the School of Computer Science and Technology, Tianjin University, Tianjin, China in 2015. 
	
	He is currently a professor with the Institute of Information Engineering, Chinese Academy of Sciences. His research interests include computer vision, multimedia, and machine learning.
\end{IEEEbiography}

\begin{IEEEbiography}[{\includegraphics[width=1in,height=1.25in,clip,keepaspectratio]{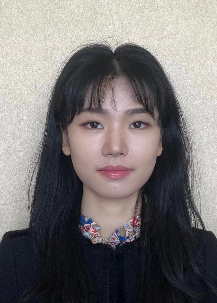}}]{Jingzhi Li} received the Ph.D. degree in cyberspace security from the University of Chinese Academy of Sciences, Beijing, China. 
	
She is currently an Associate Professor with the Institute of Information Engineering, Chinese Academy of Sciences. She was selected to join the Science and Technology Think Tank Young Talents Program of the China Association for Science and Technology. Her current research interests include image processing and face security.
\end{IEEEbiography}
%

\begin{IEEEbiography}[{\includegraphics[width=1in,height=1.25in,clip,keepaspectratio]{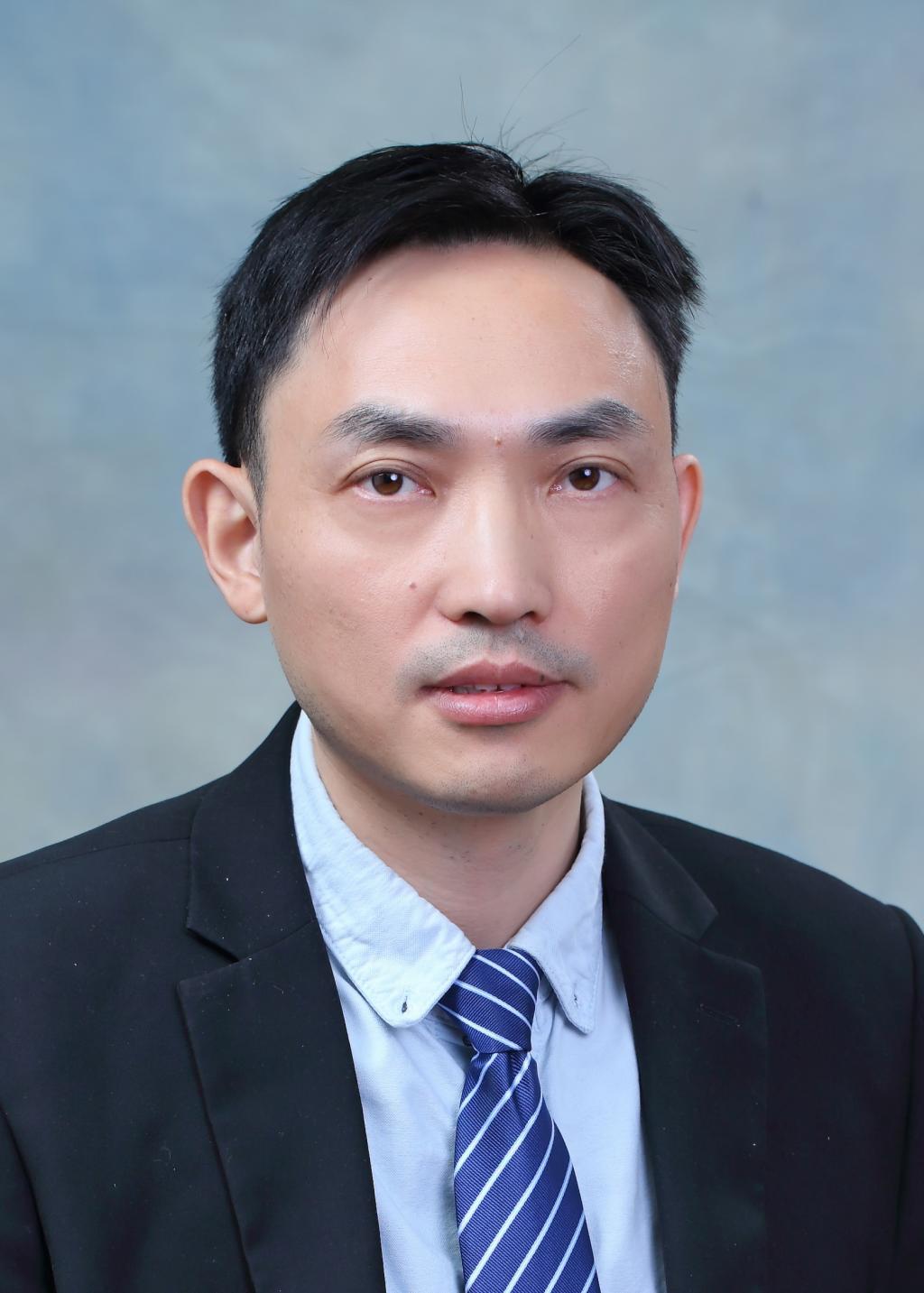}}]{Zhong Zhou} received the B.S. degree in material physics from Nanjing University in 1999 and the Ph.D. degree in computer science and technology from Beihang University, Beijing, China, in 2005. 
	
	He is currently a Professor and Ph.D. Adviser at the State Key Laboratory of Virtual Reality Technology and Systems, Beihang University. His main research interests include virtual reality, augmented reality, computer vision, and artificial intelligence.
\end{IEEEbiography}

\end{document}